\documentclass[review]{elsarticle}

\usepackage{lineno,hyperref}
\journal{Journal of \LaTeX\ Templates}

\usepackage{color}
\usepackage{amsmath}
\usepackage{amssymb}        % for ams symbols
\usepackage{graphicx}

\usepackage{amsthm}
\usepackage{lscape}
\usepackage{arydshln}
\usepackage{slashbox}
\usepackage{subcaption}

\usepackage{algorithm}
\usepackage{algorithmic}
\usepackage{multirow}

\emergencystretch=\maxdimen
\begin{document}

\begin{frontmatter}

\title{Single image super-resolution using self-optimizing mask via fractional-order gradient interpolation and reconstruction}

\author[mymainaddress,mysecondaryaddress]{Qi Yang}
%\address{}
%\fntext[myfootnote]{.}
%\ead{9501133@163.com}
\author[mymainaddress,mysecondaryaddress]{Yanzhu Zhang}

\author[mysecondaryaddress]{Tiebiao Zhao}

\author[mysecondaryaddress]{YangQuan Chen\corref{mycorrespondingauthor}}
\cortext[mycorrespondingauthor]{Corresponding author}
\ead{ychen53@ucmerced.edu}

\address[mymainaddress]{School of Mechanical Engineering, ShenYang LiGong University, Shenyang 110159, China}
\address[mysecondaryaddress]{School of Engineering, University of California, Merced. 5200 N. Lake Road, Merced, CA 95343, USA}

\begin{abstract}
Image super-resolution using self-optimizing mask via fractional-order gradient interpolation and reconstruction aims to recover detailed information from low-resolution images and reconstruct them into high-resolution images. Due to the limited amount of data and information retrieved from low-resolution images, it is difficult to restore clear, artifact-free images, while still preserving enough structure of the image such as the texture. This paper presents a new single image super-resolution method which is based on adaptive fractional-order gradient interpolation and reconstruction. The interpolated image gradient via optimal fractional-order gradient is first constructed according to the image similarity and afterwards the minimum energy function is employed to reconstruct the final high-resolution image. Fractional-order gradient based interpolation methods provide an additional degree of freedom which helps optimize the implementation quality due to the fact that an extra free parameter $\alpha$-order is being used. The proposed method is able to produce a rich texture detail while still being able to maintain structural similarity even under large zoom conditions. Experimental results show that the proposed method performs better than current single image super-resolution techniques.
\end{abstract}


\end{frontmatter}

%\linenumbers

\section{Introduction}
\begin{figure}[htbp]
  \centering
   \begin{subfigure}[b]{0.5\textwidth}
   \centering
        \includegraphics[width=2.3in]{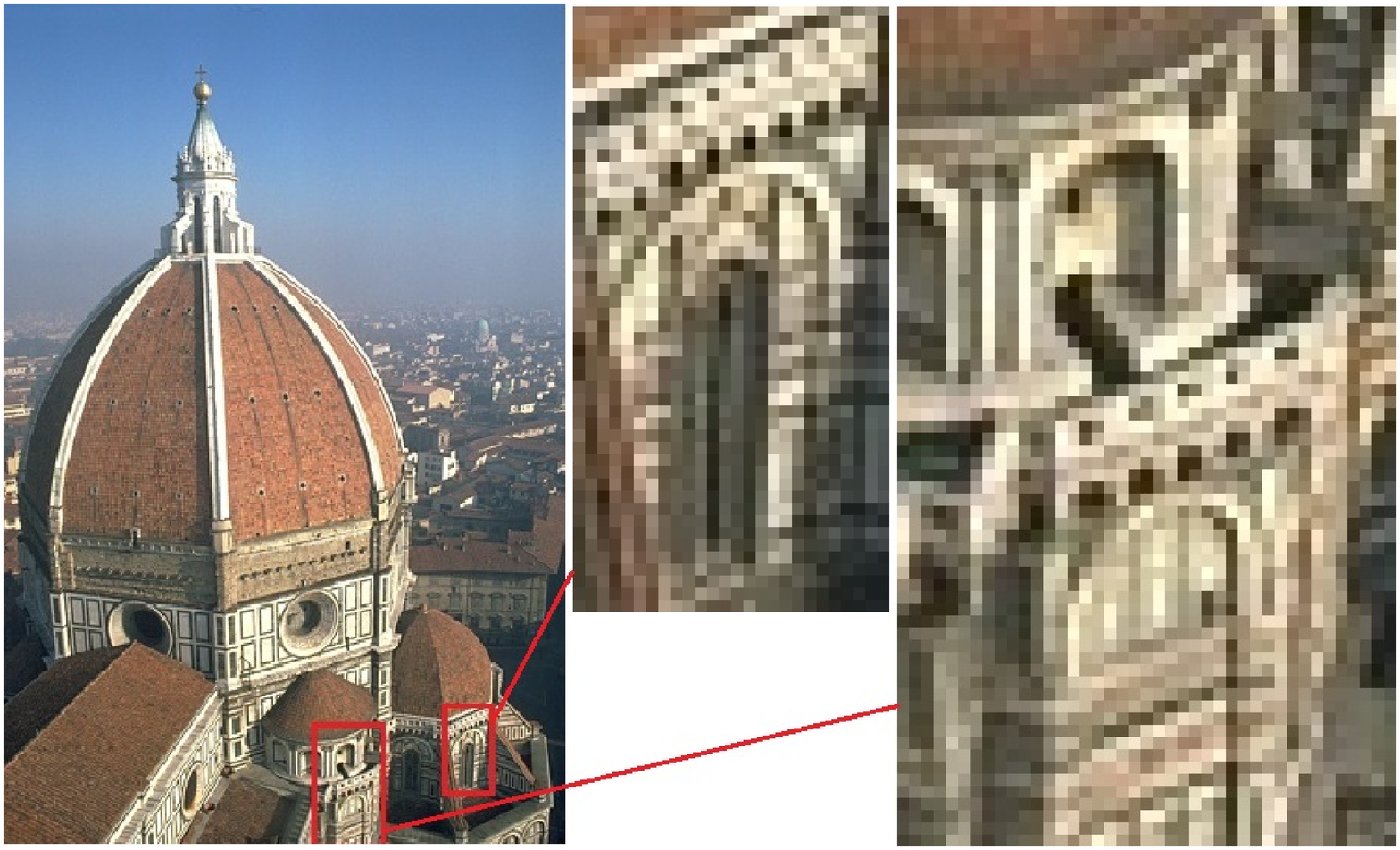}
        \caption{{\tt Low-resolution}}
    \end{subfigure}%
   \begin{subfigure}[b]{0.5\textwidth}
   \centering
        \includegraphics[width=2.3in]{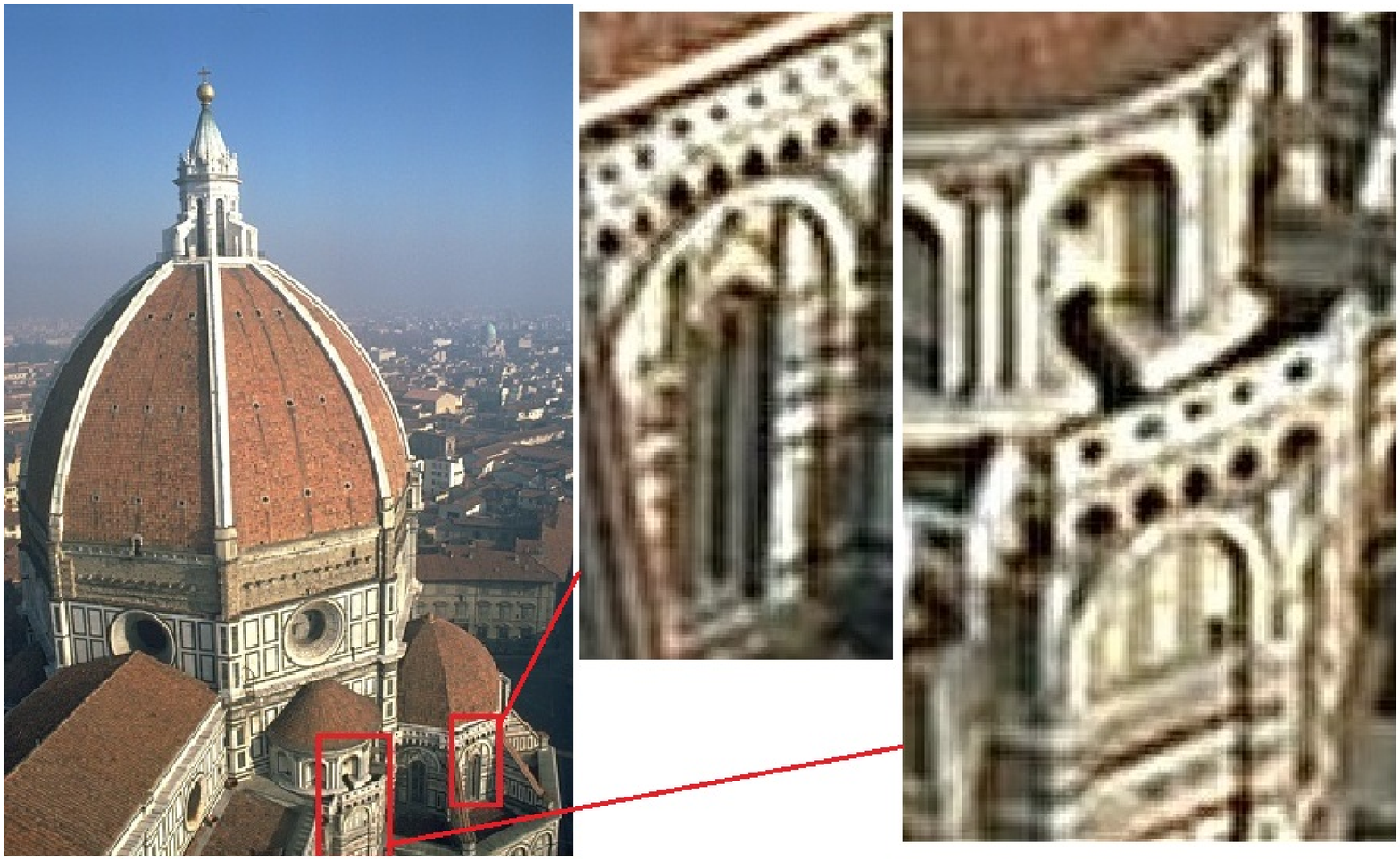}
        \caption{{\tt High-resolution}}
   \end{subfigure}%
   \caption{Our method of SR image (scale = 4). This figure is better viewed on screen with high-resolution display.}
  \label{fig:example}
\end{figure}

The goal of the single image super-resolution (SR) is to synthesize a high-resolution (HR) image with more details from only one low-resolution image. It has attracted much attention in recent years due to its variety of applications. Some of those applications include, agriculture image analysis, video surveillance and remote sensing imaging. Generally speaking, for 2D images, image SR techniques can be divided into two categories: multi-image SR and single-image SR. Multi-image SR methods mentioned in the articles \cite{boulanger2007space,farsiu2004fast,fransens2007optical} use multiple frames of the same scene to integrate one high-resolution image. Single image SR methods only use one low-resolution image for synthesis which leads to a non-unique solution. In other words, in SR methods, one low-resolution image could produce numerous high-resolution images whereas in HR methods, multiple low resolution frames of the same image could produce a single high resolution image every time. Therefore, image priors should be imposed in order to help all high-resolution images with maintain image similarity and structural similarity because image priors uses previous information about the images to enhance results. As of now, lots of methods have been proposed in single image SR techniques. These methods can be categorized into three different ways: interpolation-based methods, example-based methods and reconstruction based methods.\\
Interpolation-based methods usually utilize a base function to fit the unknown pixels in the high-resolution grids. For example, bilinear and bicubic interpolations are commonly used methods that are both simple and effective. Due to the assumption of smoothness, these methods always generate obvious visual artifacts, such as blurring and jaggedness. To handle the artifacts seen using these methods, more methods have been proposed as seen in the articles \cite{li2001new,su2004image}.\\
Example-based image SR methods however predict the desired pixels by matching the patch pairs to a universal set of training samples which can be seen in the articles, \cite{freeman2002example,tang2016example,chang2004super,timofte2013anchored,zhu2014single,zontak2011internal,zhang2010non}. This is why a big set of patches is needed to predict the correct pixels however, such a large set of patches results in an excessively heavy computational cost. However,  various methods have been proposed to help improve the calculation speed and performance. Article \cite{yang2013fast} proposes a fast regression model for practical single image super-resolution which is based on in-place examples. This example leverages two fundamental super-resolution approaches by learning from an external database and learning from self-examples. Article \cite{yang2010image} presents a new approach to single-image super-resolution which is based upon sparse signal representation. Article \cite{glasner2009super} however proposes a framework to combine the power of classical SR and example-based SR. All these approaches are based on the observation that patches in natural images tend to redundantly occur within the image, both on the same scale and across different scales. Therefore, these methods could cause artifacts due to the fact that the patch pairs being replaced are too similar to one another, especially the images that have patches of lower similarity.\\
Reconstruction-based methods obtain an SR image estimate by imposing certain prior information which tends to form global constraints in order to ensure the fidelity between the newly reconstructed high-resolution image and the original low-resolution image. This method is discussed in the following articles, \cite{zhang2015single,chen2016single,protter2009generalizing,aly2005image,ren2013fractional,li2016fractional}. Given the similarity of the structure, gradient profile prior has got quite a bit of attention recently as seen in the following articles, \cite{baker2002limits,xian2016single,sun2011gradient,fattal2007image,sun2008image}. Article \cite{xian2016single} proposes an internal gradient similarity method which produces high-resolution image gradient samples which are used for further image reconstruction. Article \cite{sun2011gradient} proposes using natural image gradients profile prior for image SR and the image gradient in this technique is modeled by using a parametric profile model. Article \cite{fattal2007image} proposes a new method for upsampling images. This new method is capable of generating sharp edges with reduced input-resolution grid-related artifacts. These reconstruction-based approaches are either focused on the sharp edge feature, or the internal or the external similarity replacement technique, both of which increases the ambiguity among images. This tends to synthesize the detail texture as an unnatural style. Furthermore, when the image has a discontinuous gradient or a non-smooth gradient, the methods will ultimately cause a higher deviation.\\
In this work, we present a unique and effective single image SR method using self-optimizing fractional-order gradient interpolation and reconstruction. The high-resolution multi-scale image is created based on a linear spatial pyramid. The layer image of the pyramid is first constructed according to the image similarity using the optimal fractional order gradient interpolation method, and then the minimum energy function is employed to reconstruct the final high-resolution layer image. Our presented method combines the advantage of adaptive fractional-order gradient interpolation approaches and the reconstruction-based techniques. Self-optimizing fractional-order gradient interpolation approaches preserve enough image structure such as the texture, while the reconstruction-based techniques ensure clear, artifacts-free images. As shown in Fig. \ref{fig:example}, the comparison of the experimental results shows that our approach not only successfully restores clear edges, but also restores enough of the texture aspect of the structure. Due to the page size limit, this figure is better viewed on a screen with a higher resolution display. Compared to other single image SR methods, our present approach has the following advantages:\\
%\textcolor[rgb]{0,0,1}{
(i) The proposed adaptive fractional-order gradient interpolation method provides an additional degree of freedom (parameter $\alpha$ -order) in optimizing implementation quality which ensures that we are able to produce high-resolution image while maintaining clear textures and edges.\\
(ii) The minimum energy function is used to ensure the global fidelity between the low-resolution image and the high-resolution image.\\
(iii) Linear spatial pyramid structure upsampling provides the multi-scale fidelity constraint.\\%}
The remainder of this paper is organized as follows: in Section \ref{sec:Linear spatial pyramid and framework}, we introduce the linear spatial pyramid and its framework. In section \ref{sec:Adaptive fractional-order gradient interpolation} we illustrate the self-optimizing fractional-order gradient interpolation algorithm. In the Section \ref{sec:Reconstruct high-resolution image}, we introduce the high-resolution image construction method. In section \ref{sec:Experimental results} we look at a comparison of image and texture similarity. In Section \ref{sec:Conclusion} we will conclude the paper with a summary of our findings.

\section{Linear spatial pyramid and framework}
\label{sec:Linear spatial pyramid and framework}
\begin{figure}[h]
\centering
\includegraphics[width=0.7\textwidth]{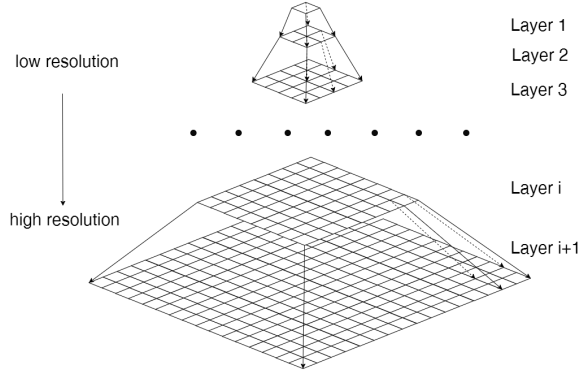}
\caption{The structure of SR linear spatial pyramid}
\label{fig:The structure of SR pyramid}
\end{figure}

\begin{figure}[h]
\centering
\includegraphics[width=0.5\textwidth]{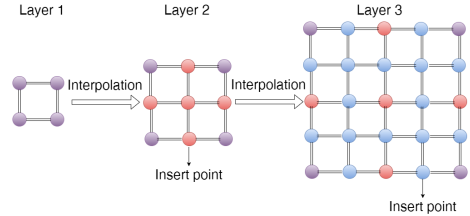}
\caption{The layer of SR linear spatial pyramid interpolation method}
\label{fig:The layer of SR linear spatial pyramid interpolation method}
\end{figure}

\begin{figure}[h]
\centering
\includegraphics[width=1.0\textwidth]{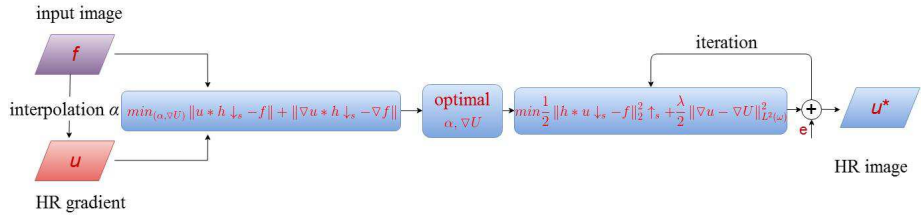}
\caption{The framework of proposed method in each layer of pyramid}
\label{fig:The framework of proposed method}
\end{figure}

A pyramid is a pattern of multi-layer or multi-scale signal representation which is mainly used in image processing and computer vision. In a pyramid, an image is repeatedly either upsampling or downsampling. Pyramid representation is a predecessor to the multi-resolution operation. We use the SR linear spatial pyramid to construct the high-resolution images because it offers a multi-scale operation feature. Compared with other linear interpolation approaches, our SR linear spatial pyramid method which uses the multi-layer minimum energy function is able to ensure fidelity between the input low-resolution image and the output high-resolution image. The structure of the pyramid is shown in Fig. \ref{fig:The structure of SR pyramid}.
Given the certain level of the pyramid, we are able to utilize the interpolation method. The value of the inserted point is calculated through the fractional-order gradient value which is optimized by tuning the $\alpha$-order parameter. The layer of SR linear spatial pyramid interpolation method is shown in Fig. \ref{fig:The layer of SR linear spatial pyramid interpolation method}. From the figure we can see that for any four points of layer \emph{i}, we will insert five points in the middle. For the next layer $i+1$, the inserted points will be regard as known points for iterative interpolation approach. The scale factor \emph{s} is $2,4,8\cdots 2^{n}$. In the picture, the purple four points are the original points, the five red points are the first inserted points and the sixteen blue points are the second inserted points.
After establishing the SR linear spatial pyramid interpolation architecture, we will consider how to calculate the value of the inserted points in order to ensure that the synthesized high-resolution image of each layer has a strong texture feature in its structure and has excellent fidelity. The calculated process under SR linear spatial pyramid structure is shown in Fig. \ref{fig:The framework of proposed method}. The input low-frequency image is denoted as $f\in R^{N_{1}\times N_{2}}$, from which we can obtain its high-resolution image $u\in R^{sN_{1}\times sN_{2}}$ by fractional-order $\alpha$ gradient interpolation. Once we obtain the low-frequency band image by using a low-pass Gaussian filter \emph{h}, we are now able to calculate its deviation with the input image \emph{f} and search for the minimum the deviation corresponding gradient $\triangledown U$ by tuning the parameter $\alpha$-order. Once we obtain the optimal parameter $\alpha$-order and corresponding gradient $\triangledown U$ of the image, we put it to the minimum energy function (the third blue frame) to generate its final high-resolution image by using the gradient descent algorithm. The fractional-order gradient interpolation and the minimum energy function reconstruction will be discussed shortly.
\section{Self-optimizing mask via fractional-order gradient interpolation}
\label{sec:Adaptive fractional-order gradient interpolation}
\begin{figure}[h]
\centering
\includegraphics[width=0.5\textwidth]{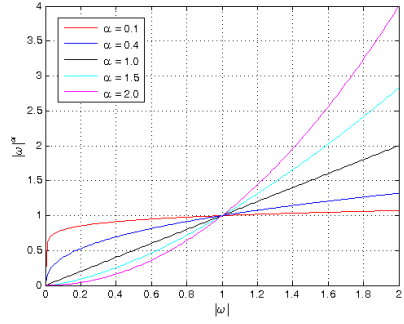}
\caption{Amplitude-frequency curves of fractional-order}
\label{fig:Amplitude-frequency curves of fractional-order}
\end{figure}

\begin{figure}[h]
\centering
   \begin{subfigure}[b]{0.34\textwidth}
   \centering
        \includegraphics[width=1.6in]{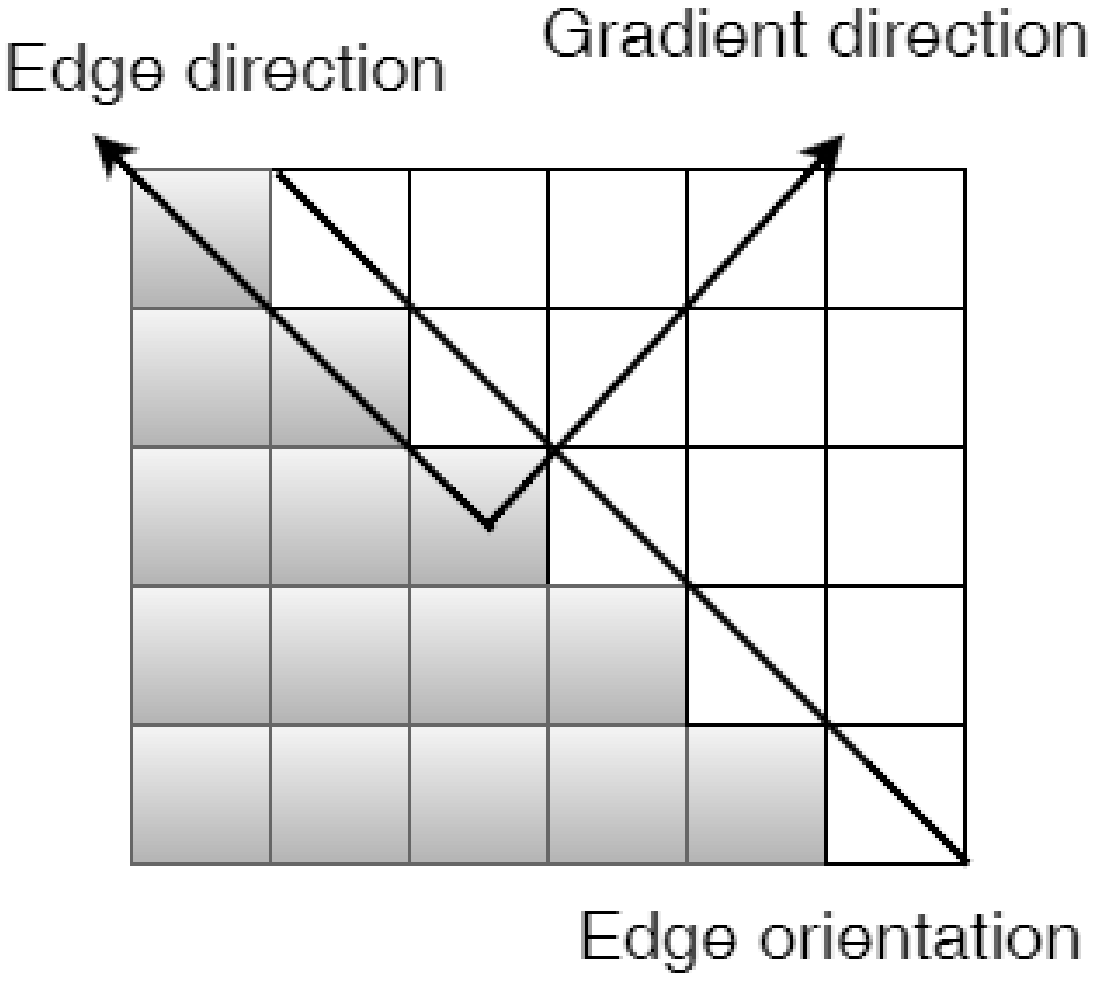}
        \caption{{\tt Center point}}
    \end{subfigure}%
\centering
   \begin{subfigure}[b]{0.34\textwidth}
   \centering
        \includegraphics[width=1.6in]{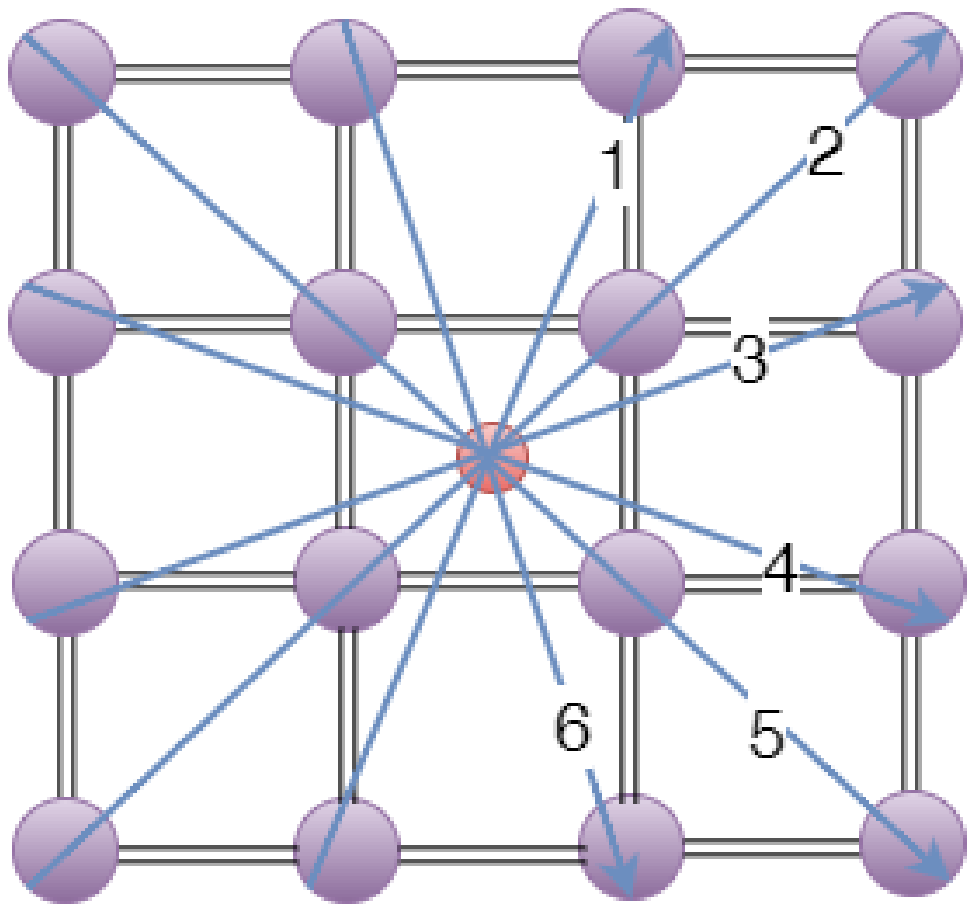}
        \caption{{\tt Center point}}
    \end{subfigure}%
   \begin{subfigure}[b]{0.34\textwidth}
   \centering
        \includegraphics[height=1.6in]{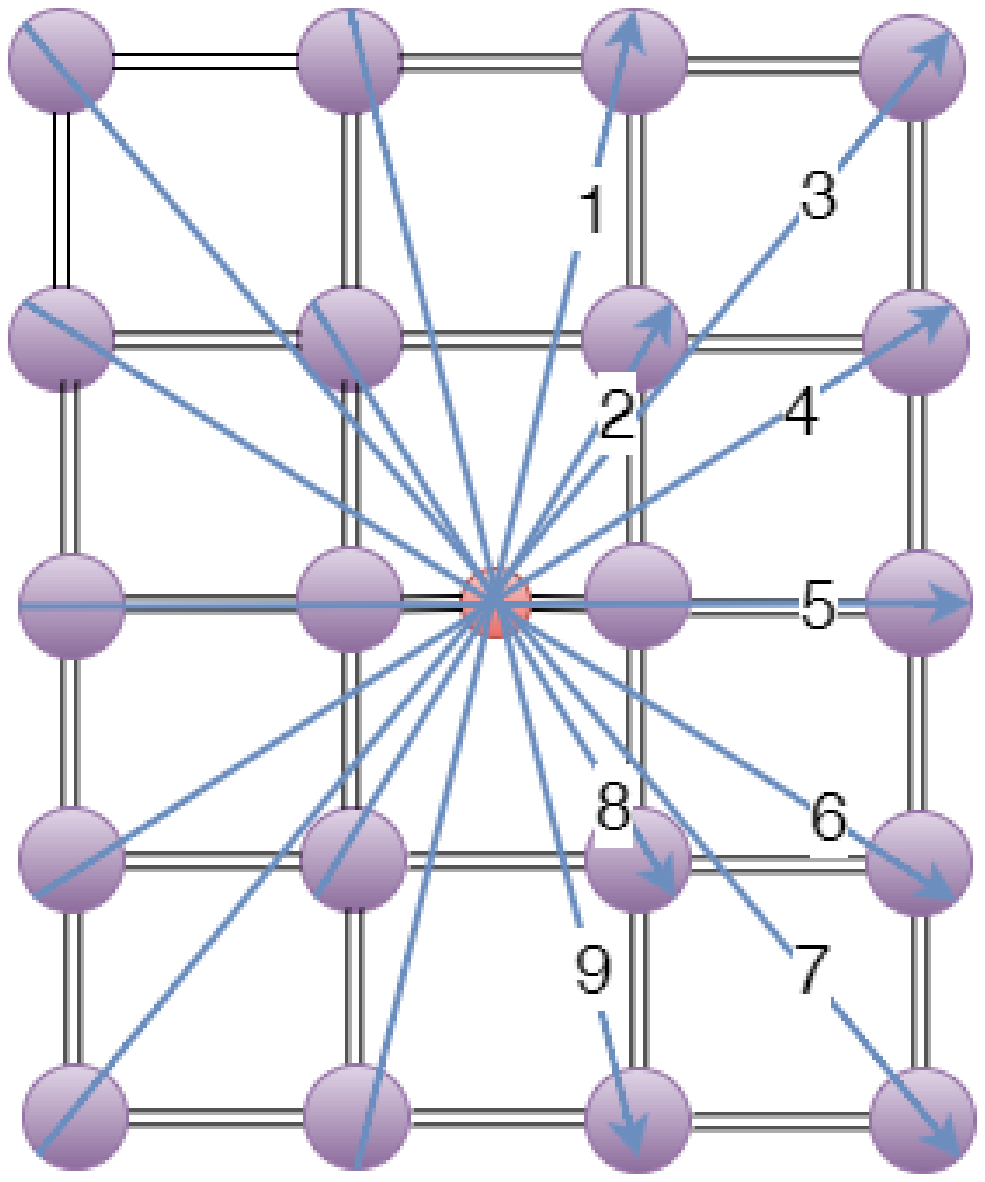}
        \caption{{\tt Between point}}
   \end{subfigure}%
\caption{Point interpolation}
\label{fig:Center point interpolation}
\end{figure}

\begin{figure}[htbp]
  \centering
   \begin{subfigure}[b]{0.17\textwidth}
   \centering
        \includegraphics[width=0.7in]{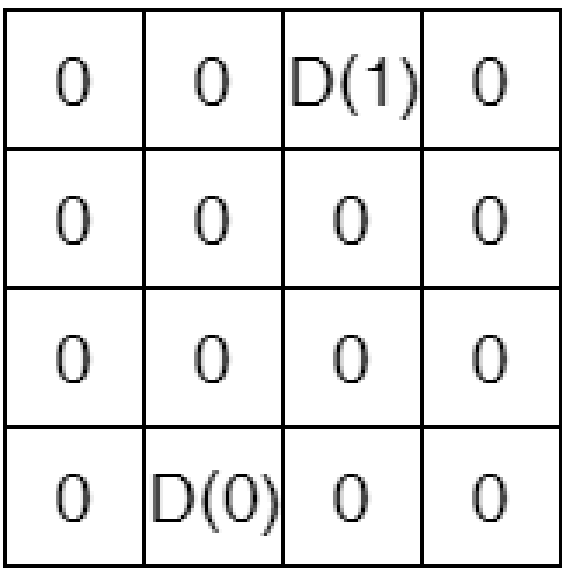}
        \caption{{Direction 1}}
    \end{subfigure}%
   \begin{subfigure}[b]{0.17\textwidth}
   \centering
        \includegraphics[width=0.7in]{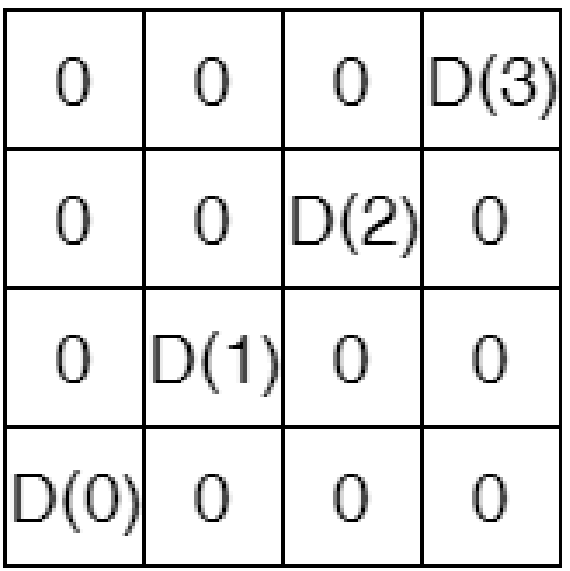}
        \caption{{Direction 2}}
   \end{subfigure}%
   \begin{subfigure}[b]{0.17\textwidth}
   \centering
        \includegraphics[width=0.7in]{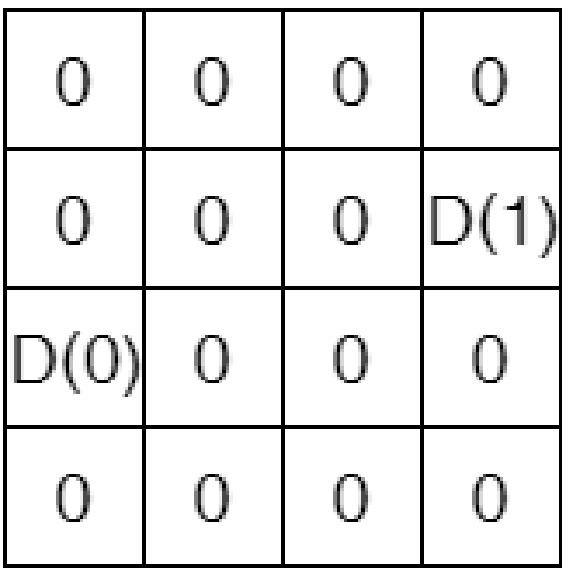}
        \caption{{Direction 3}}
    \end{subfigure}%
   \begin{subfigure}[b]{0.17\textwidth}
   \centering
        \includegraphics[width=0.7in]{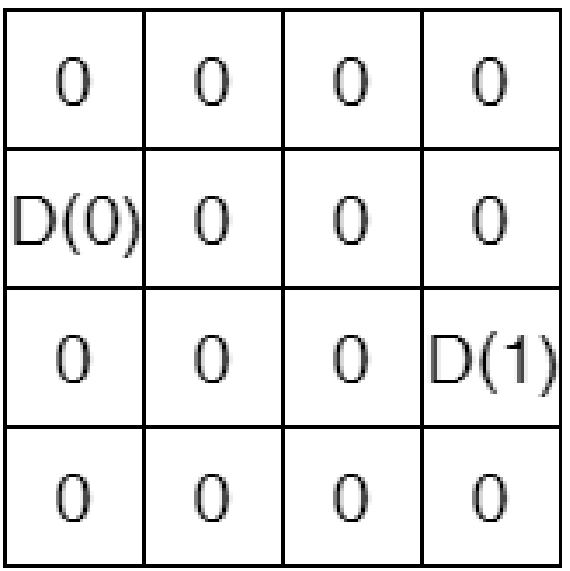}
        \caption{{Direction 4}}
   \end{subfigure}%
   \centering
    \begin{subfigure}[b]{0.17\textwidth}
   \centering
        \includegraphics[width=0.7in]{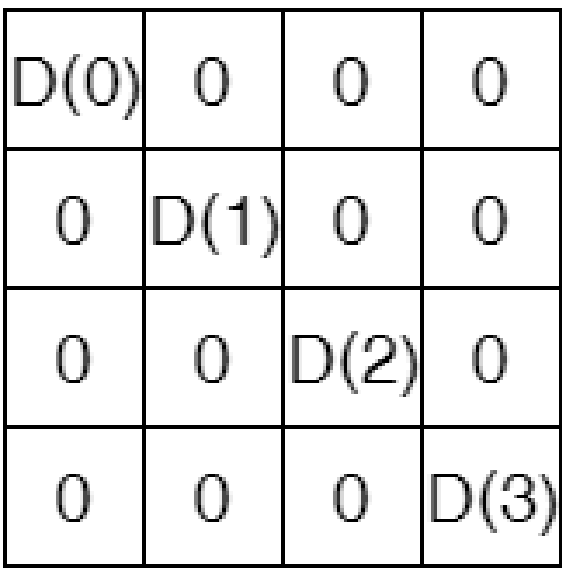}
        \caption{{Direction 5}}
    \end{subfigure}%
   \begin{subfigure}[b]{0.17\textwidth}
   \centering
        \includegraphics[width=0.7in]{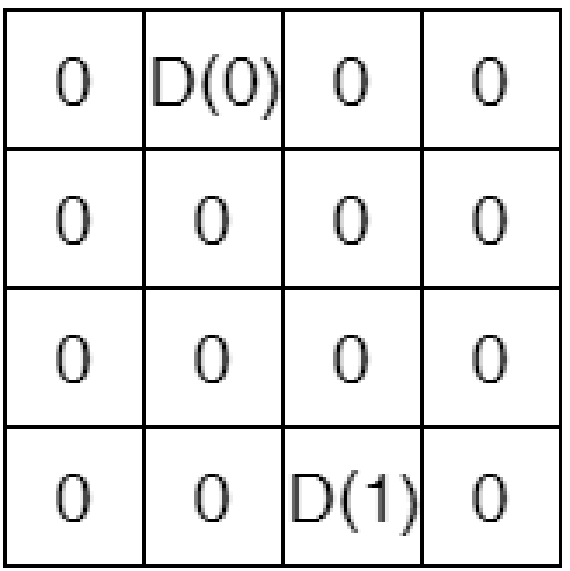}
        \caption{{Direction 6}}
   \end{subfigure}%
    \caption{Six directions mask}
    \label{fig:Six direction result}
\end{figure}

The subject of fractional calculus and its applications have gained considerable popularity during the past few decades or so in various fields of science and engineering, such as image processing. For image texture and edge preservation, the most integral differential operators work well when used for high-frequency features of images such as Sobel, Prewitt, and Laplacian of Gaussian operators. However, their performance deteriorates significantly when applied to smooth regions. While the fractional differential operator has the capability of not only preserving high-frequency contour features, it also has the ability to improve the low-frequency texture details in smooth areas as seen in Fig. \ref{fig:Amplitude-frequency curves of fractional-order}. Based on this idea, we utilize the fractional-order gradient interpolation method in order to maintain the clear texture and preserve the edges of high-resolution images.\\
Three popular definitions for fractional calculus were given by Gr\"{u}nwald Letnikov (G-L), Riemann Liouville (R-L), and Caputo. Of these, G-L is the most popular definition used in digital image processing as explained in article \cite{yang2016fractional}. For numerical calculation of fractional-order derivatives we can use the relation derived from the G-L definition. It is given by:
\begin{equation}\label{eq-1}
_{a}^{ }\textrm{D}_{t}^{\alpha }f(t)=\lim_{h \to 0}\frac{1}{h^{\alpha }}\sum_{j=0}^{[\frac{t-a}{h}]}(-1)^{j}\binom{\alpha }{j}f(t-jh)\approx \frac{1}{h^{\alpha }}\sum_{j=0}^{[\frac{t-a}{h}]}\omega _{j}^{(\alpha )}f(t-jh).
\end{equation}
where $\omega_{j}^{(\alpha )}=(-1)^{j}\binom{\alpha }{j}$ is the polynomial coefficients of $(1-z)^{\alpha }$ which can be calculated by the following the recurrence formula \cite{dingyu2006control,chen2002discretization}:
\begin{equation}\label{eq-2}
\omega _{0}^{(\alpha )}=1,\; \omega _{j}^{(\alpha )}=\left ( 1-\frac{\alpha +1}{j} \right )\omega _{j-1}^{(\alpha )},\; j=1,2,...
\end{equation}
As for the 2D image, the duration of the image is divided by equal intervals $h = 1$ and $t\in \left [ b-a \right ]$. Therefore the $\alpha$-order fractional derivatives of $f(t)$ is approximated by:
\begin{equation}\label{eq-3}
{ }\textrm{D}_{t}^{\alpha }f(t)\approx f(t)+(-\alpha )f(t-1)+\frac{(-\alpha )(-\alpha +1)}{2}f(t-2)+\cdots +\frac{\Gamma (-\alpha +1)}{n!\Gamma (-\alpha -n+1)}f(t-n).
\end{equation}
The backward difference of the fractional partial derivatives on the negative $x$ and $y$ are expressed as follows \cite{li2015adaptive}:
\begin{equation}\label{eq-4}
\begin{split}
{ }\textrm{D}_{x}^{\alpha }f(x,y)\approx f(x,y)+(-\alpha )f(x-1,y)+\frac{(-\alpha )(-\alpha +1)}{2}f(x-2,y)\\+\cdots +\frac{\Gamma (-\alpha +1)}{n!\Gamma (-\alpha -n+1)}f(x-n,y).
\end{split}
\end{equation}
\begin{equation}\label{eq-5}
\begin{split}
{ }\textrm{D}_{y}^{\alpha }f(x,y)\approx f(x,y)+(-\alpha )f(x,y-1)+\frac{(-\alpha )(-\alpha +1)}{2}f(x,y-2)\\+\cdots +\frac{\Gamma (-\alpha +1)}{n!\Gamma (-\alpha -n+1)}f(x,y-n).
\end{split}
\end{equation}
Based on formula \ref{eq-4} and formula \ref{eq-5}, the $\alpha$th-order derivative of the middle point in the mask window can be estimated by formula \ref{eq-6}, where $D(n)=\frac{\Gamma (-\alpha +1)}{n!\Gamma (-\alpha -n+1)}$ is the coefficients of the mask:
\begin{equation}\label{eq-6}
D_{n}^{\alpha }=\left [ D(0),\cdots ,D(n) \right ].
\end{equation}
To enable smoothness of the interpolated image, we employ the fractional-order gradient mask of the edge direction (tangent to the contour direction) in order to calculate the value of the interpolation point, which can be seen in Fig. \ref{fig:Center point interpolation}. Edge direction of the input image patch is obtained from the following equation \ref{eq-6-1}:
\begin{equation}\label{eq-6-1}
\theta = 90 +  tan^{-1}(\frac{\triangledown f_{y}}{\triangledown f_{x}}).
\end{equation}
where $\triangledown f_{x}$, $\triangledown f_{y}$ and $\theta$ refer to the the gradient in the $x$ direction, the the gradient in $y$ direction and the edge direction, respectively. As for center point of interpolation, we construct six direction masks, which are shown in Fig. \ref{fig:Six direction result}. For between point interpolations, we construct nine direction masks, which are shown in following Fig. \ref{fig:Nine direction result}. We select the mask based on the nearest angle.
\begin{figure}[htbp]
  \centering
   \begin{subfigure}[b]{0.2\textwidth}
   \centering
        \includegraphics[width=0.7in]{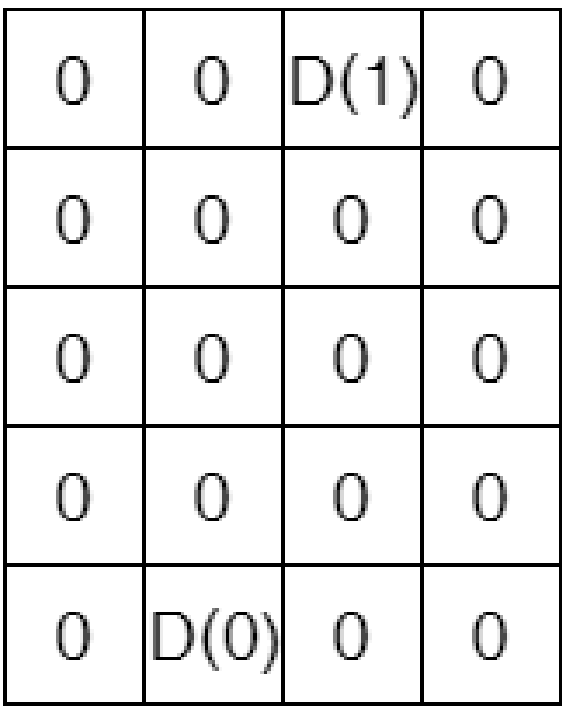}
        \caption{{Direction 1}}
    \end{subfigure}%
   \begin{subfigure}[b]{0.2\textwidth}
   \centering
        \includegraphics[width=0.7in]{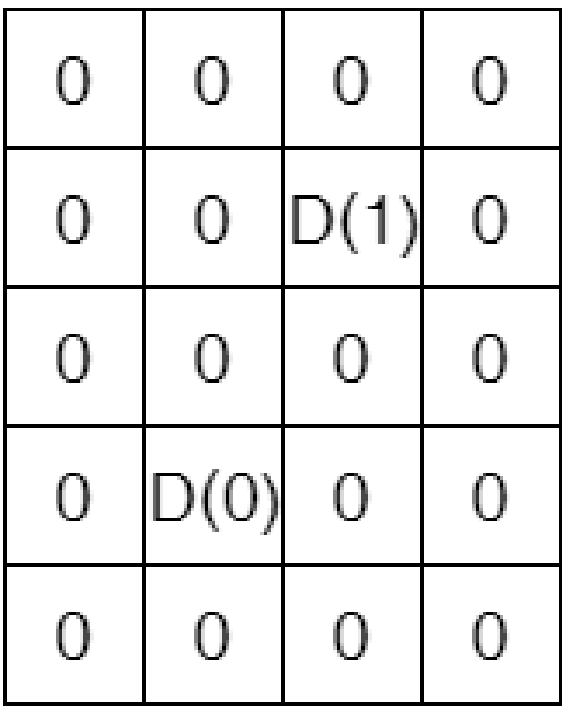}
        \caption{{Direction 2}}
   \end{subfigure}%
   \centering
   \begin{subfigure}[b]{0.2\textwidth}
   \centering
        \includegraphics[width=0.7in]{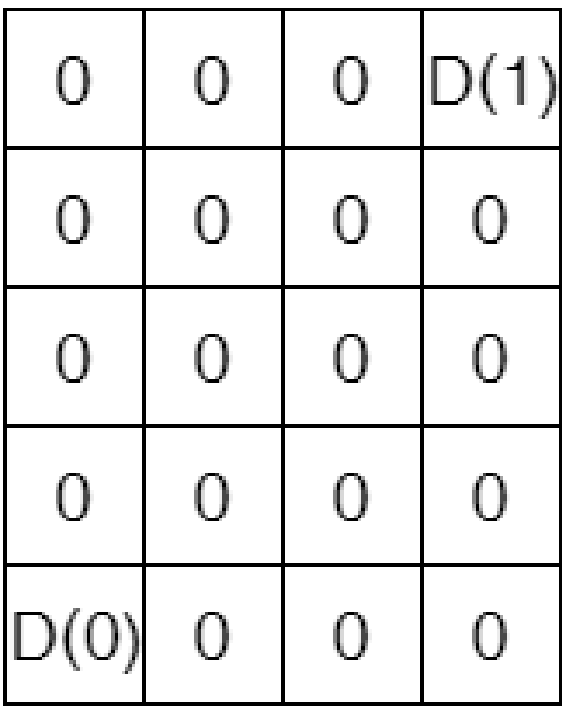}
        \caption{{Direction 3}}
    \end{subfigure}%
   \begin{subfigure}[b]{0.2\textwidth}
   \centering
        \includegraphics[width=0.7in]{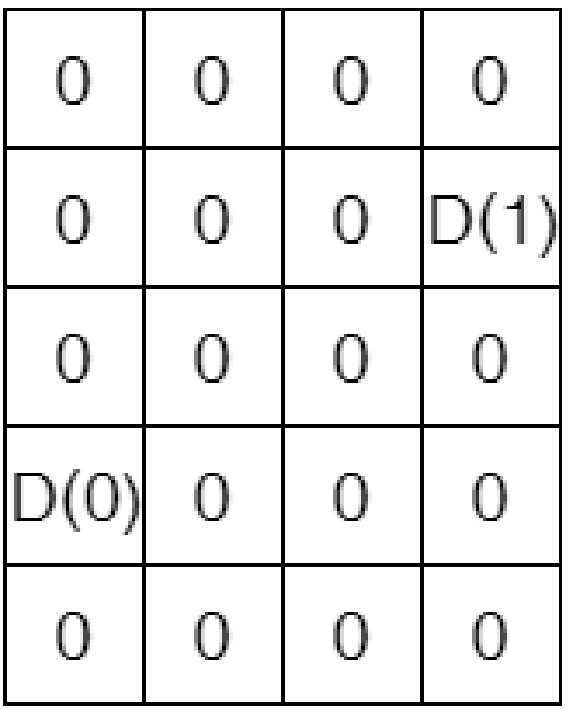}
        \caption{{Direction 4}}
   \end{subfigure}%
   \\
   \begin{subfigure}[b]{0.2\textwidth}
   \centering
        \includegraphics[width=0.7in]{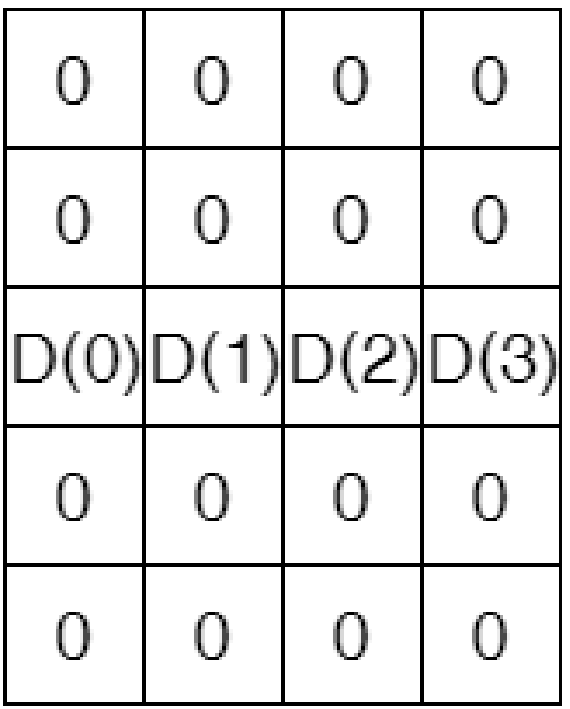}
        \caption{{Direction 5}}
    \end{subfigure}%
   \begin{subfigure}[b]{0.2\textwidth}
   \centering
        \includegraphics[width=0.7in]{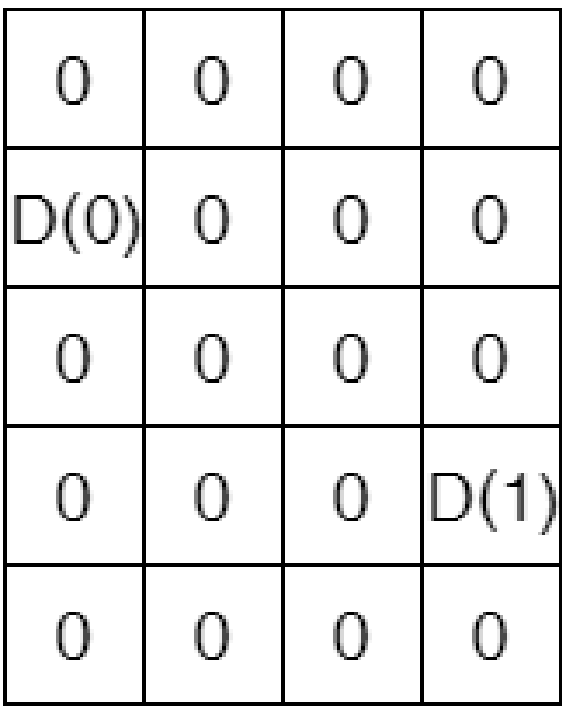}
        \caption{{Direction 6}}
   \end{subfigure}%
   \begin{subfigure}[b]{0.2\textwidth}
   \centering
        \includegraphics[width=0.7in]{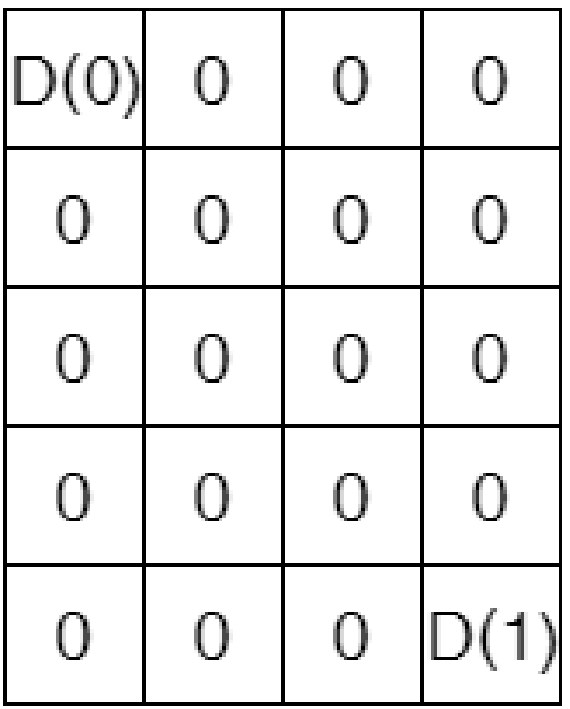}
        \caption{{Direction 7}}
    \end{subfigure}%
   \begin{subfigure}[b]{0.2\textwidth}
   \centering
        \includegraphics[width=0.7in]{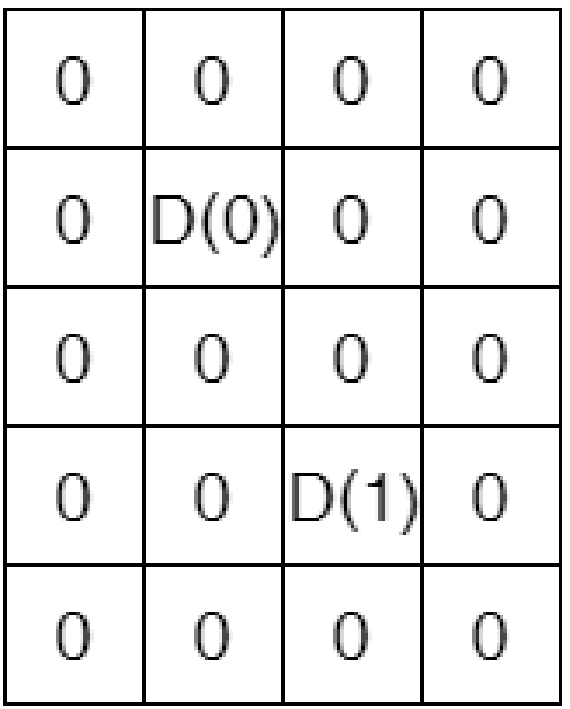}
        \caption{{Direction 8}}
   \end{subfigure}%
   \begin{subfigure}[b]{0.2\textwidth}
   \centering
        \includegraphics[width=0.7in]{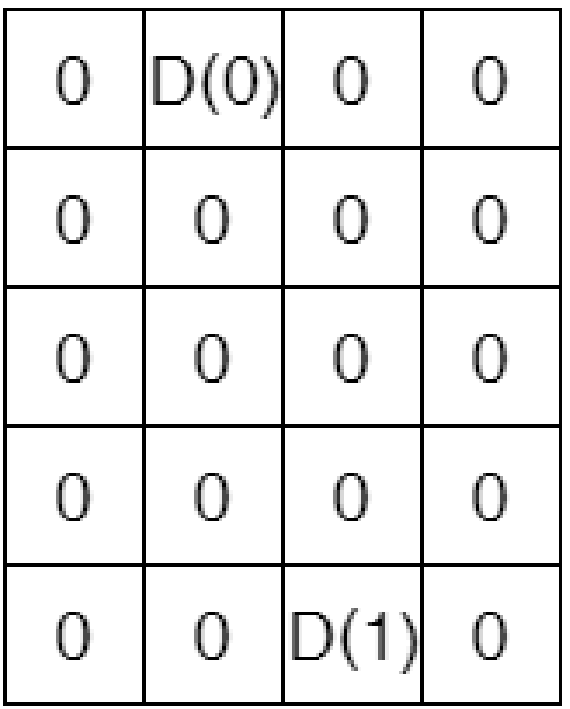}
        \caption{{Direction 9}}
   \end{subfigure}%
    \caption{Nine directions mask}
    \label{fig:Nine direction result}
\end{figure}
Given the low-resolution image patch $ f_{x,y}$ and its nearest mask $D_{\theta }^{\alpha }$, the gradient of the inserted points is denoted as $\triangledown u_{x,y}$.
%The new points follow the linear interpolation method.
It can be seen in the following formula:
\begin{equation}\label{eq-7}
\triangledown u_{x,y} =  f_{x,y}\times D_{\theta }^{\alpha }.
\end{equation}

\begin{equation}\label{eq-77}
u_{x,y} =  f_{x,y}\pm \triangledown u_{x,y}
\end{equation}

In order to ensure the fidelity between the input image and the interpolation image, the fidelity term is imposed in order to get the optimal parameter $\alpha$ as well as gradient $\triangledown U$ by tuning the parameter $\alpha$:
\begin{equation}\label{eq-8}
min_{(\alpha ,\triangledown U)}\left \|  u*h\downarrow_{s}-  f\right \|+\left \| \triangledown u*h\downarrow_{s}- \triangledown f\right \|.
\end{equation}
\section{Reconstruct high-resolution image}
\label{sec:Reconstruct high-resolution image}
With the optimal $\triangledown U$ and input image $f$, the final high-resolution image is reconstructed using the minimum energy function that is shown in formula \ref{eq-9}:
\begin{equation}\label{eq-9}
C(u) = \frac{1}{2}\left \| h*u\downarrow_{s}-f \right \|_{2}^{2}\uparrow_{s} +\frac{\lambda }{2}\left \| \triangledown u-\triangledown U \right \|_{L^{2}(\omega )}^{2}.
\end{equation}
where $u$ is the interpolation image and $h$ stands for the Gaussian kernel with a mean of value zero and standard variance $\sigma$. The variance $\sigma$ changes with different scaling factors $s$, usually as 0.55. Operator $*$ represents the convolution operation. Operator $\downarrow$ is the downsampling operator and Operator $\uparrow$ is the upsampling operator. The coefficient of terms $\lambda $ is assigned as 0.05 to balance the fidelity term and gradient similarity term. The gradient descent optimization algorithm is used for the final high-resolution image as explained in article \cite{zeiler2012adadelta}. Instead of accumulating the sum of squared gradients over all time, we limit the window of past gradients that are accumulated and then we multiply those past gradients using some fixed values. This ensures that the learning step increases when the error is large, and the learning step decreases when the error is small. Assume at time $t$ this average is $sum\left [ g^{2} \right ]_{t}$, we are then able to calculate it using formulas \ref{eq-10} - \ref{eq-14}. The parameter $\eta $ is assigned as 1.5 to accelerate the iteration. The algorithmic process is shown as follows:
\begin{equation}\label{eq-10}
g_{t}=\frac{\partial C(u)}{\partial u}=h^{T}*(h*u\downarrow_{s}-f)\uparrow_{s}+\lambda \triangledown^{T} *(\triangledown u-\triangledown U).
\end{equation}
\begin{equation}\label{eq-11}
sum\left [ g^{2} \right ]_{t} =\beta sum\left [ g^{2} \right ]_{t-1}+\gamma g_{t}^{2}.
\end{equation}
\begin{equation}\label{eq-12}
mean\left [ g \right ]_{t}=\sqrt{sum\left [ g^{2} \right ]_{t}+e}.
\end{equation}
\begin{equation}\label{eq-13}
\Delta u_{t}=-\eta \frac{mean\left [ \Delta u \right ]_{t-1}}{mean\left [ g \right ]_{t}}g_{t}.
\end{equation}
\begin{equation}\label{eq-14}
u_{t+1} = u_{t}+\Delta u_{t}.
\end{equation}
\begin{algorithm}
\caption{Optimized gradient descent\cite{zeiler2012adadelta}}
\label{alg1}
\begin{algorithmic}
\REQUIRE $\beta $, $\gamma $, $e$, initial $u_{1}$, $sum\left [ g^{2} \right ]_{0}=0,sum\left [ \Delta x^{2} \right ]_{0}=0$
\ENSURE $u_{t+1} = u_{t}+\Delta u_{t}$
\FOR {$t = 1$ to T}
     \STATE compute gradient: $g_{t}$
     \STATE accumulate gradient: $sum\left [ g^{2} \right ]_{t} =\beta sum\left [ g^{2} \right ]_{t-1}+\gamma g_{t}^{2}$
     \STATE compute updates: $\Delta u_{t}=-\eta \frac{mean\left [ \Delta u \right ]_{t-1}}{mean\left [ g \right ]_{t}}g_{t}$
     \STATE accumulate updates: $sum\left [ \Delta x^{2} \right ]_{t} =\beta sum\left [ \Delta x^{2} \right ]_{t-1}+\gamma \Delta x_{t}^{2}$
     \STATE update u: $u_{t+1} = u_{t}+\Delta u_{t}$
\ENDFOR
\RETURN $u_{t+1}$
\end{algorithmic}
\end{algorithm}
\section{Experimental results}
\label{sec:Experimental results}
In this section, we evaluate our methods with both synthetic test examples used in the super-resolution literature and test examples in the Berkeley dataset BSDS500. In both cases, our approaches show remarkable performance. As we all know the RGB color model is the most popular color model in image processing. However, it always causes color distortion due to the \lq R\rq, \lq G\rq  and \lq B\rq channels correlation as explained in article \cite{chen20121}. Because Lab and YUV models have weak coupling in its channels, we use those models instead of the RGB color model in order to keep the color stability. The process is shown as follows. First, we change the RGB color model to the Lab or YUV model, and then we only use the \lq L\rq or \lq Y\rq component to implement our present method due to the fact that it closely resembles human perception. The other channels are obtained using the bicubic algorithm. Finally, we transform it back to the RGB color model.
\subsection{Parameter selection}
Because we use the linear spatial pyramid structure, so the scale factor \emph{s} can only be to $2,4,8\cdots 2^{n}$. We iteratively calculate the scale factor in each layer of the pyramid. When using the gradient descent algorithm, the parameter $\beta =0.9$ and the parameter $\gamma =0.01$. A set of high-resolution images are generated in datasets BSDS500 as seen in article \cite{MartinFTM01} with tuning $\alpha$ between 0 and 1.
\subsection{Visual results}
\begin{figure}[htbp]
  \centering
   \begin{subfigure}[b]{0.17\textwidth}
   \centering
        \includegraphics[width=0.75in]{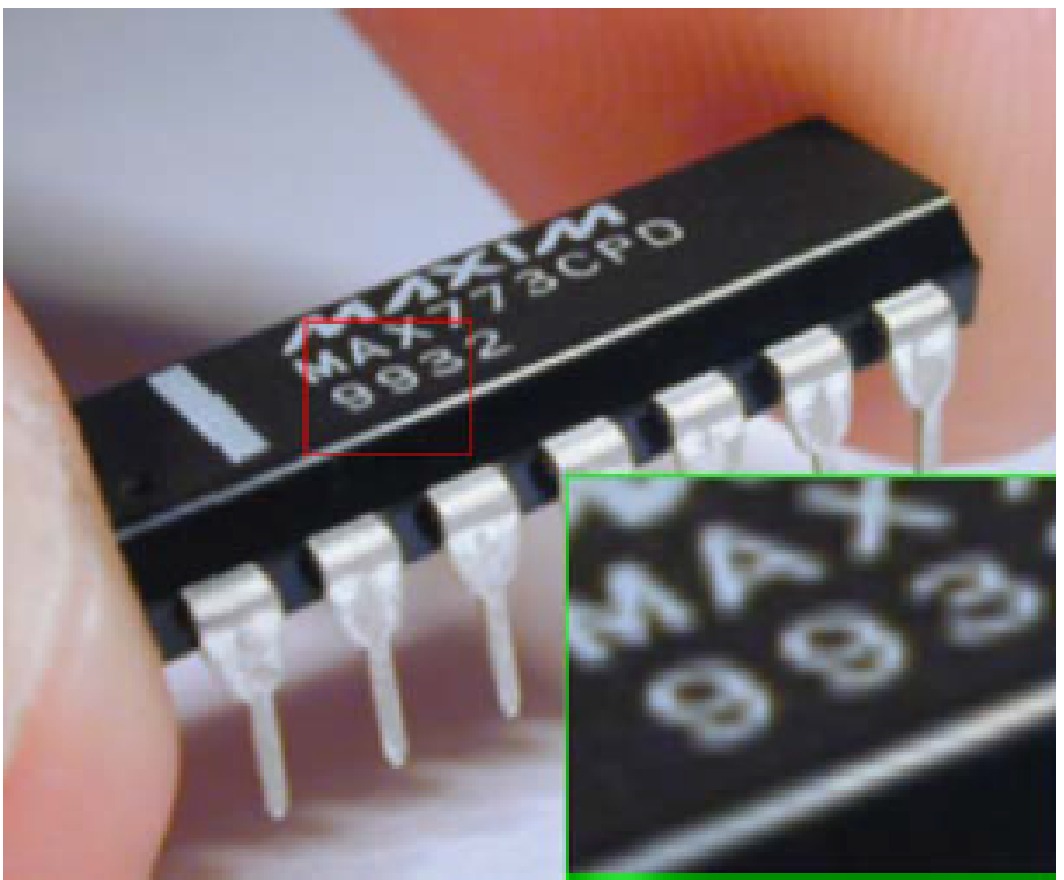}

    \end{subfigure}%
   \begin{subfigure}[b]{0.17\textwidth}
   \centering
        \includegraphics[width=0.75in]{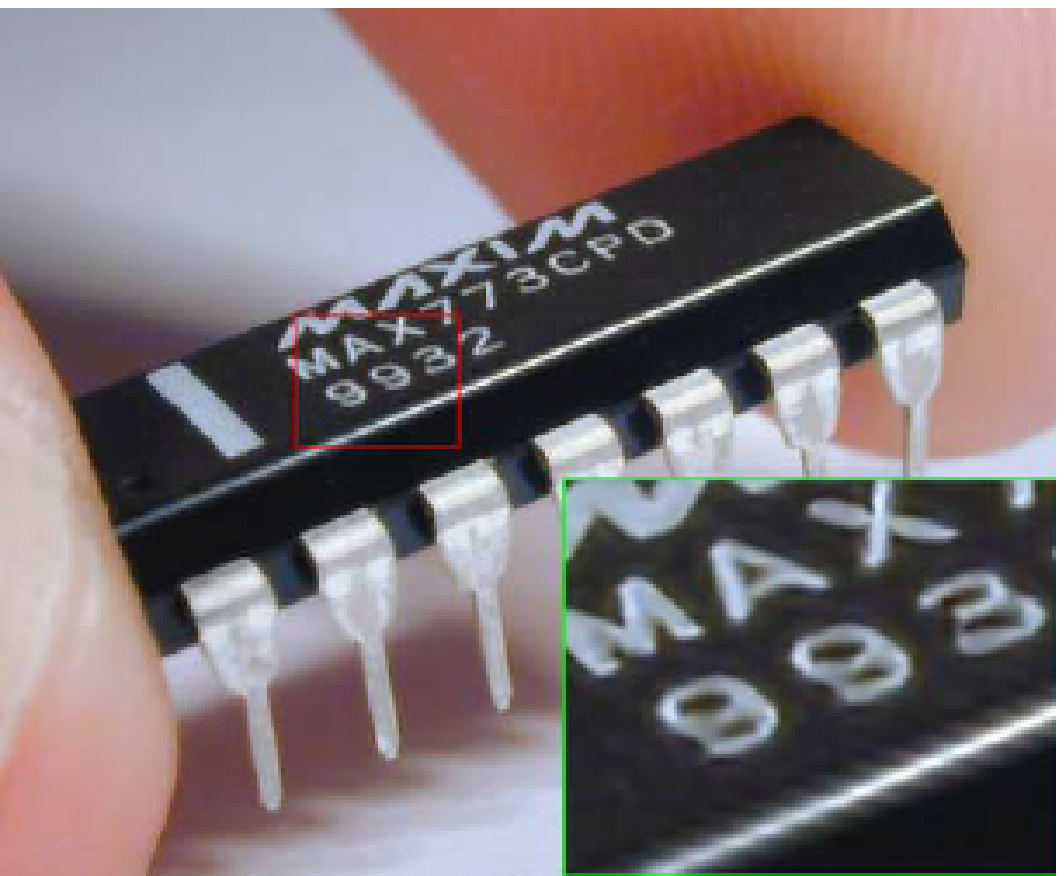}

   \end{subfigure}%
     \centering
   \begin{subfigure}[b]{0.17\textwidth}
   \centering
        \includegraphics[width=0.75in]{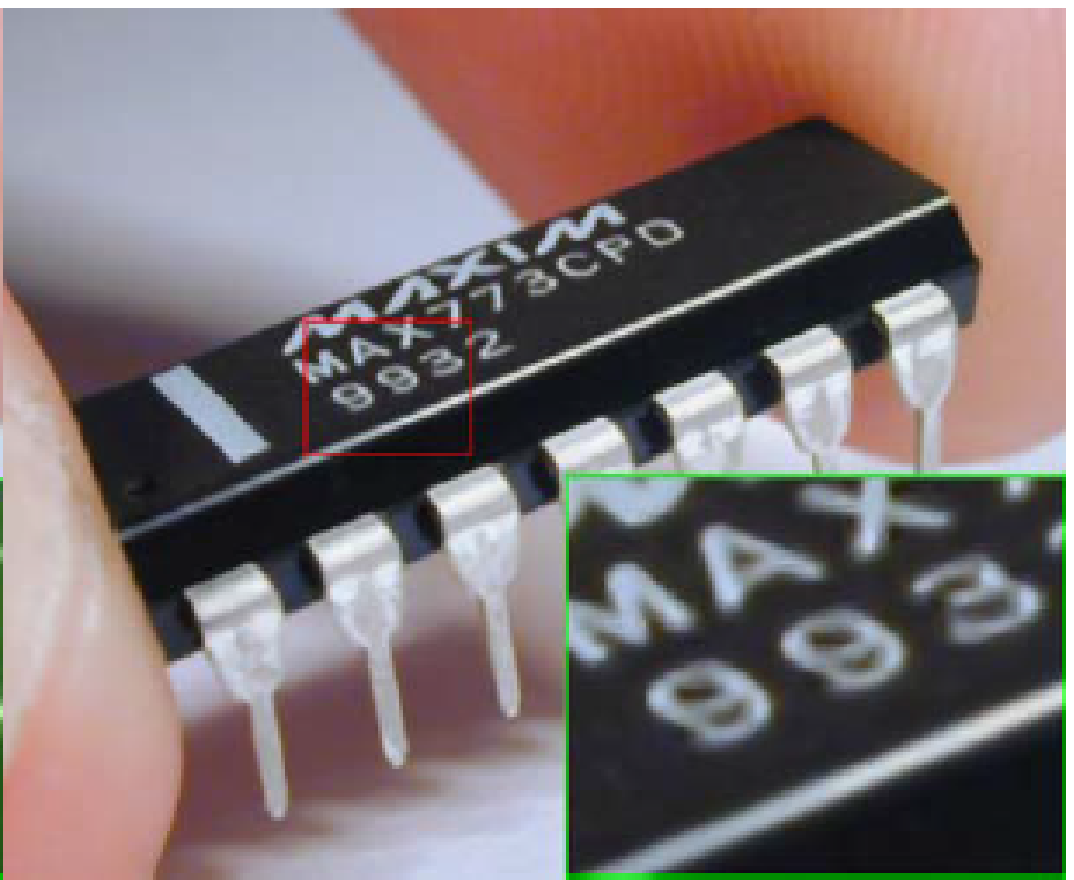}

    \end{subfigure}%
   \begin{subfigure}[b]{0.17\textwidth}
   \centering
        \includegraphics[width=0.75in]{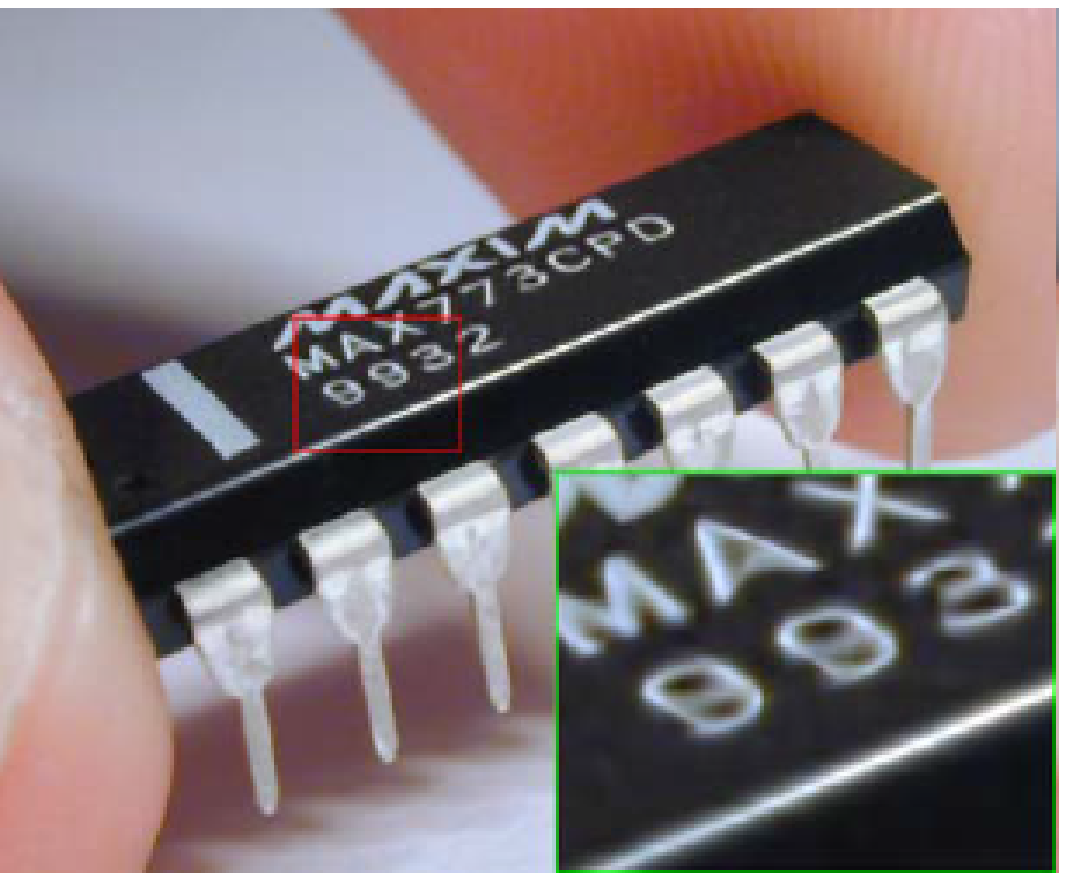}

   \end{subfigure}%
     \centering
   \begin{subfigure}[b]{0.17\textwidth}
   \centering
        \includegraphics[width=0.75in]{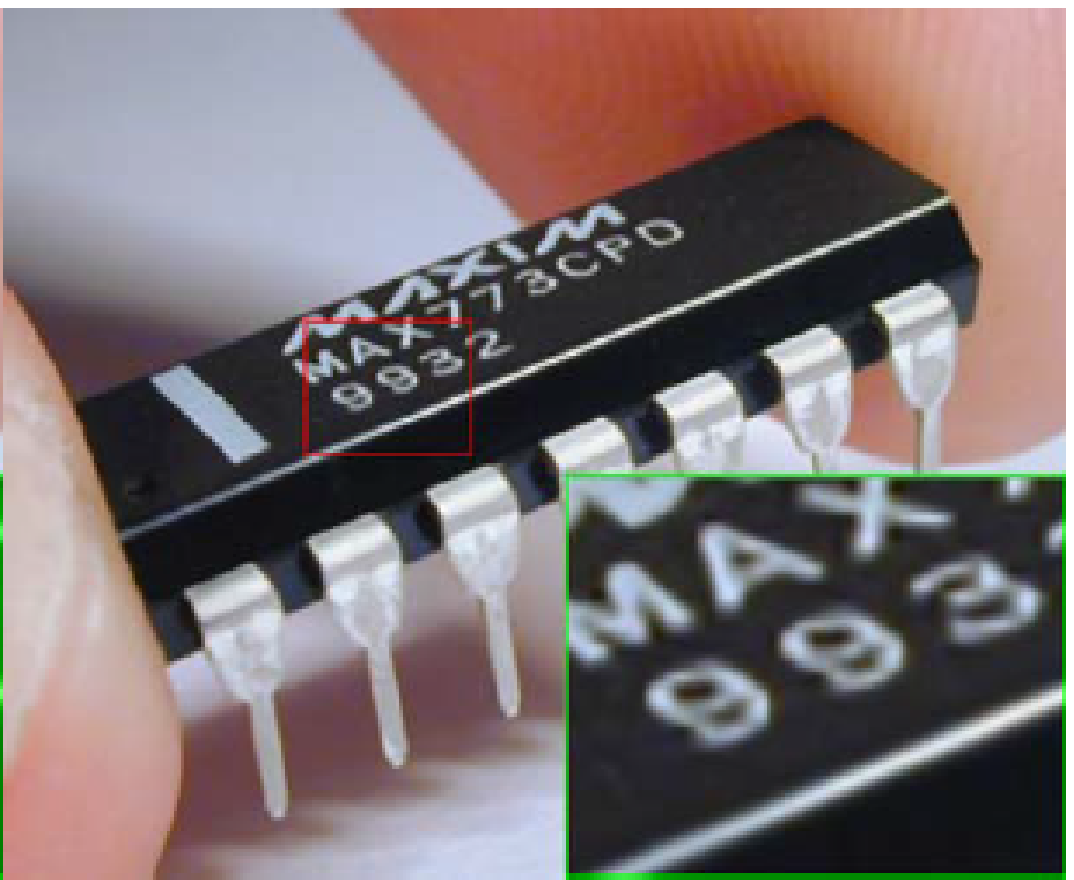}
    \end{subfigure}%
   \begin{subfigure}[b]{0.17\textwidth}
   \centering
        \includegraphics[width=0.75in]{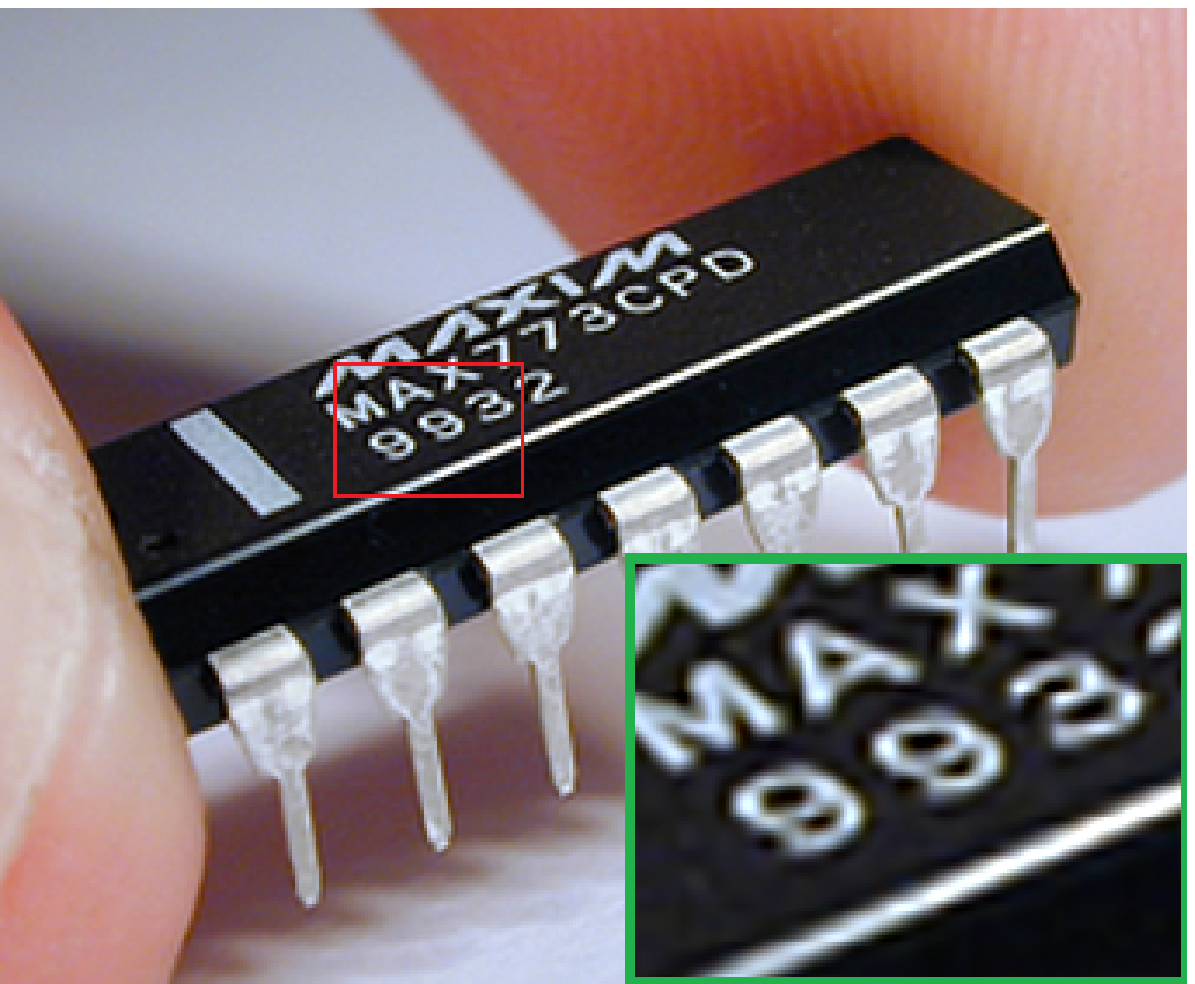}
   \end{subfigure}%
   \\
   \begin{subfigure}[b]{0.17\textwidth}
   \centering
        \includegraphics[width=0.75in]{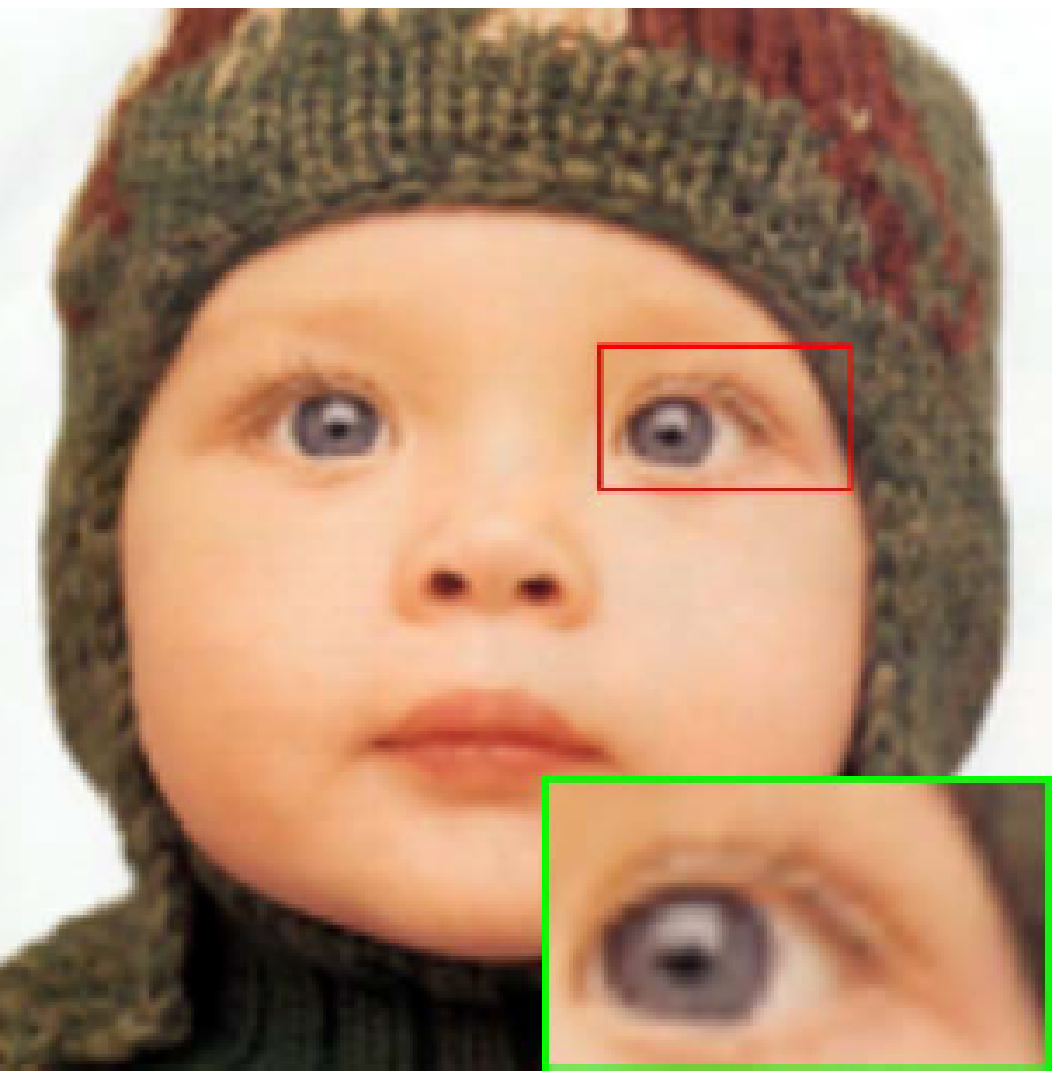}
         \caption{{Bicubic}}
    \end{subfigure}%
   \begin{subfigure}[b]{0.17\textwidth}
   \centering
        \includegraphics[width=0.75in]{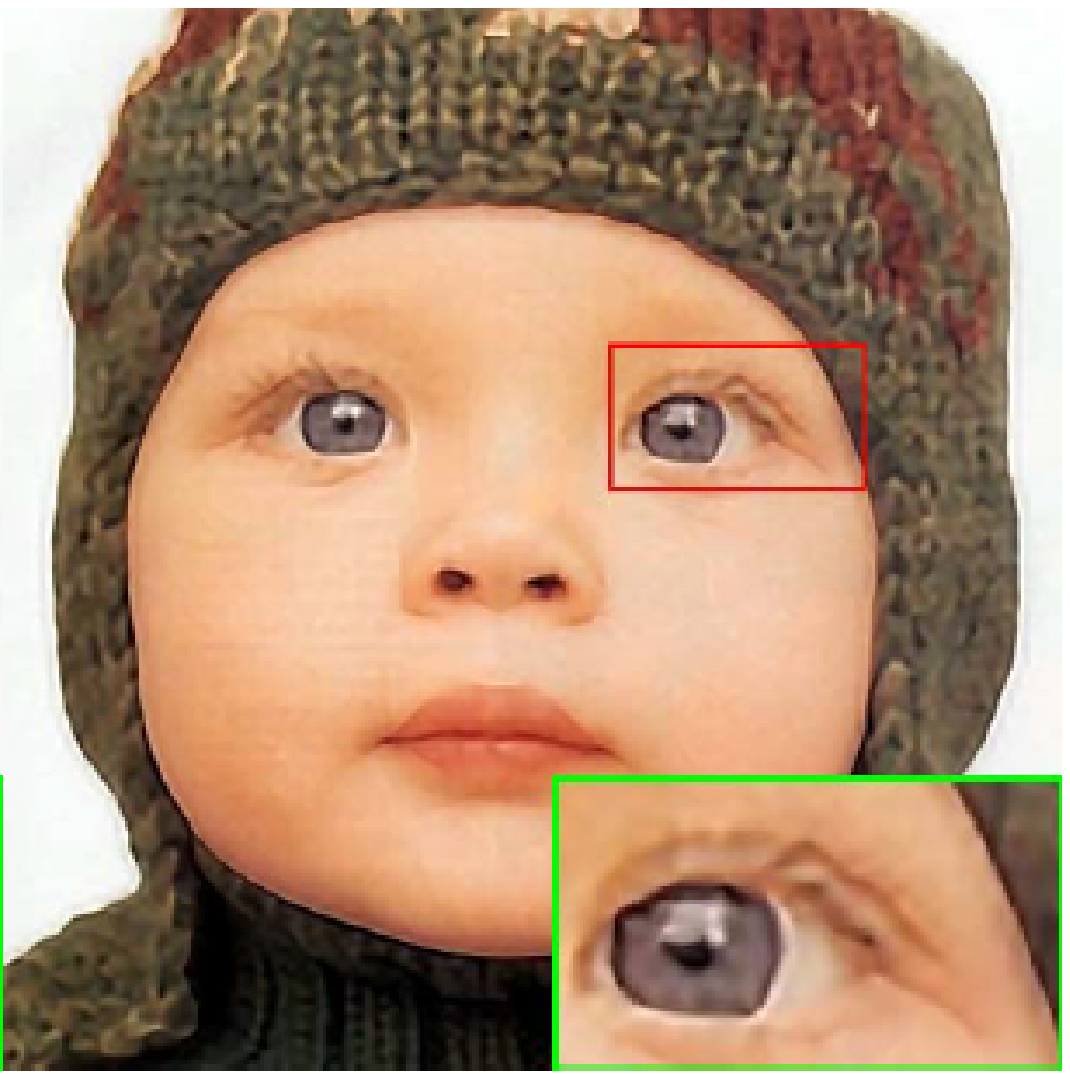}
        \caption{{\cite{glasner2009super}}}
   \end{subfigure}%
     \centering
   \begin{subfigure}[b]{0.17\textwidth}
   \centering
        \includegraphics[width=0.75in]{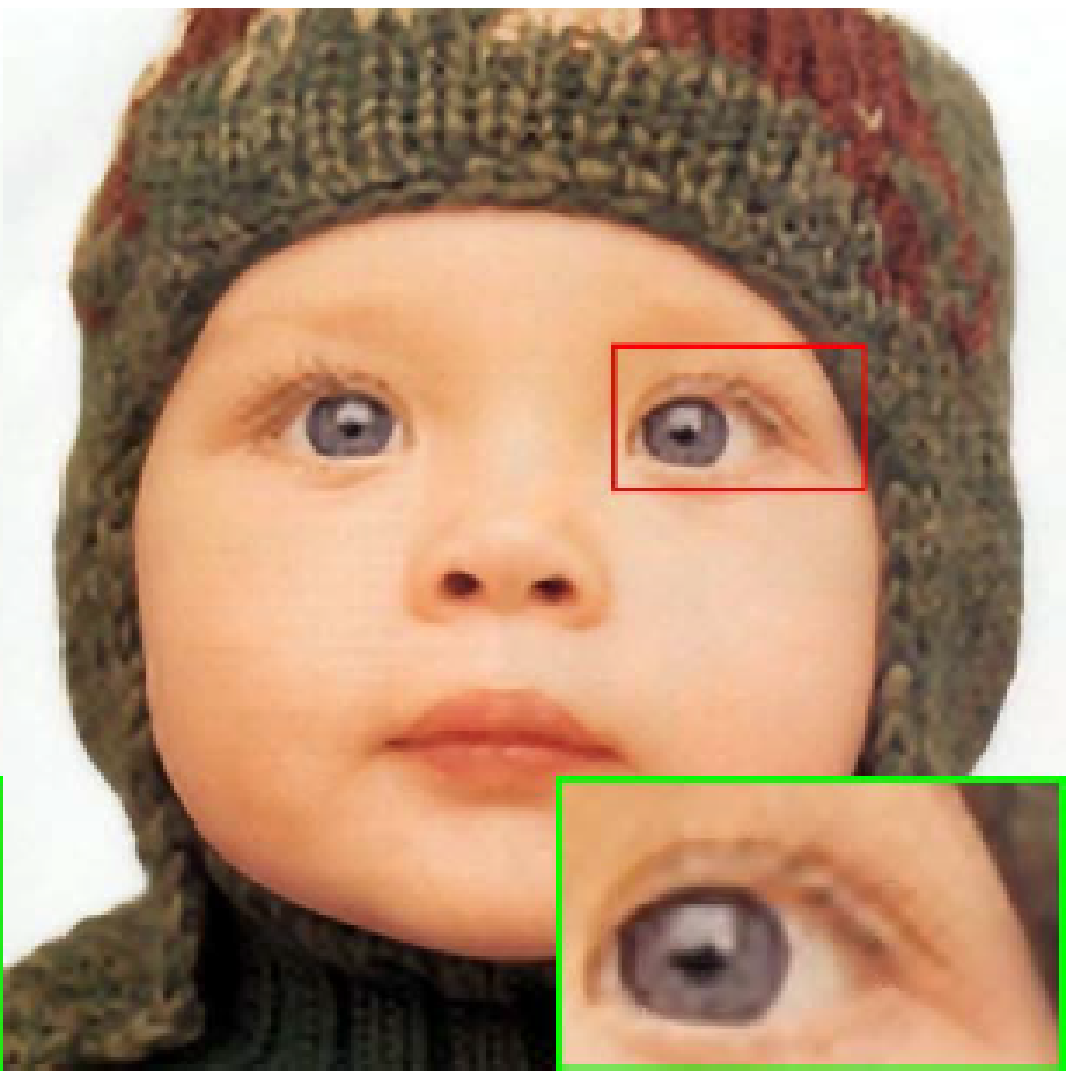}
        \caption{{\cite{yang2012coupled}}}
    \end{subfigure}%
   \begin{subfigure}[b]{0.17\textwidth}
   \centering
        \includegraphics[width=0.75in]{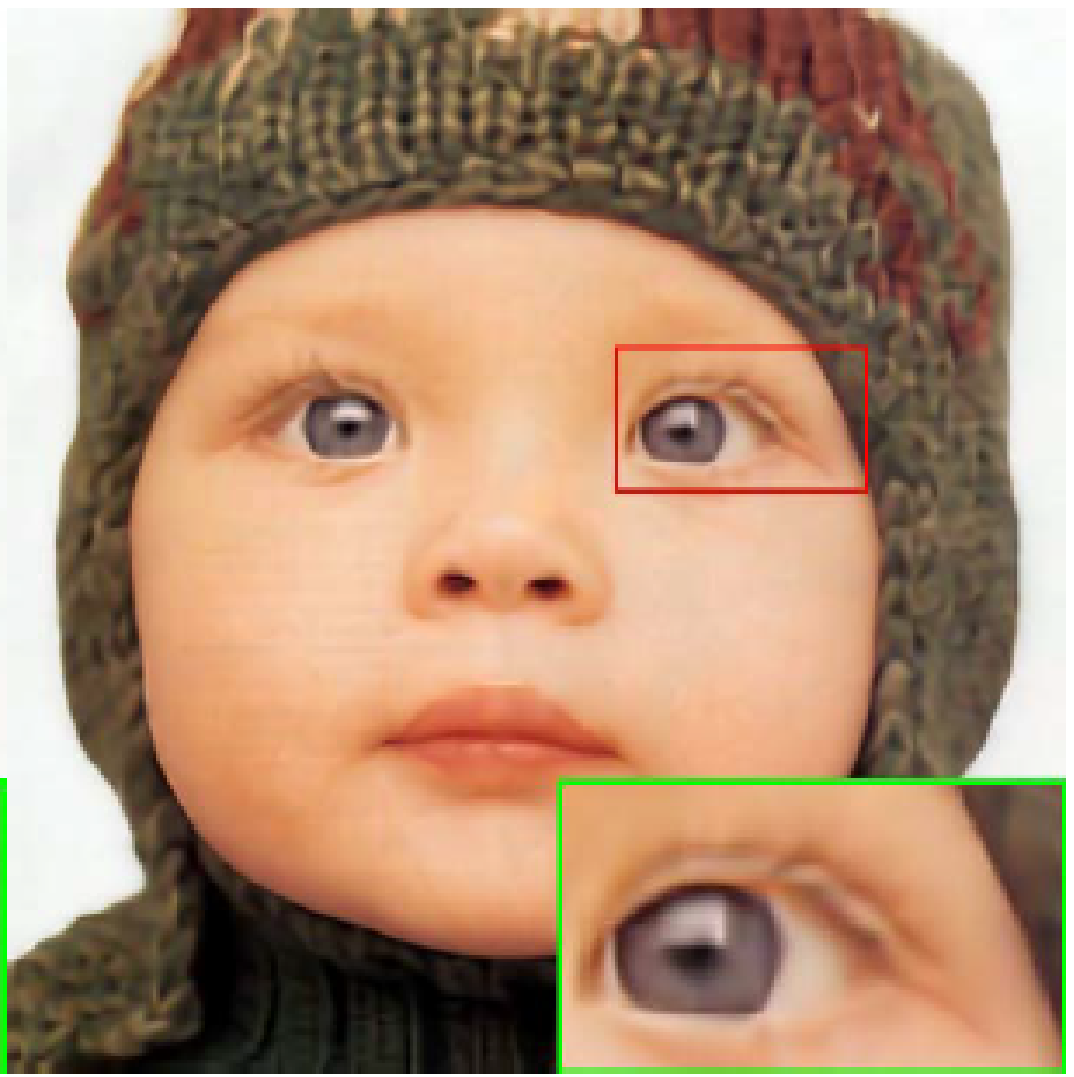}
       \caption{{\cite{freedman2011image}}}
   \end{subfigure}%
     \centering
   \begin{subfigure}[b]{0.17\textwidth}
   \centering
        \includegraphics[width=0.75in]{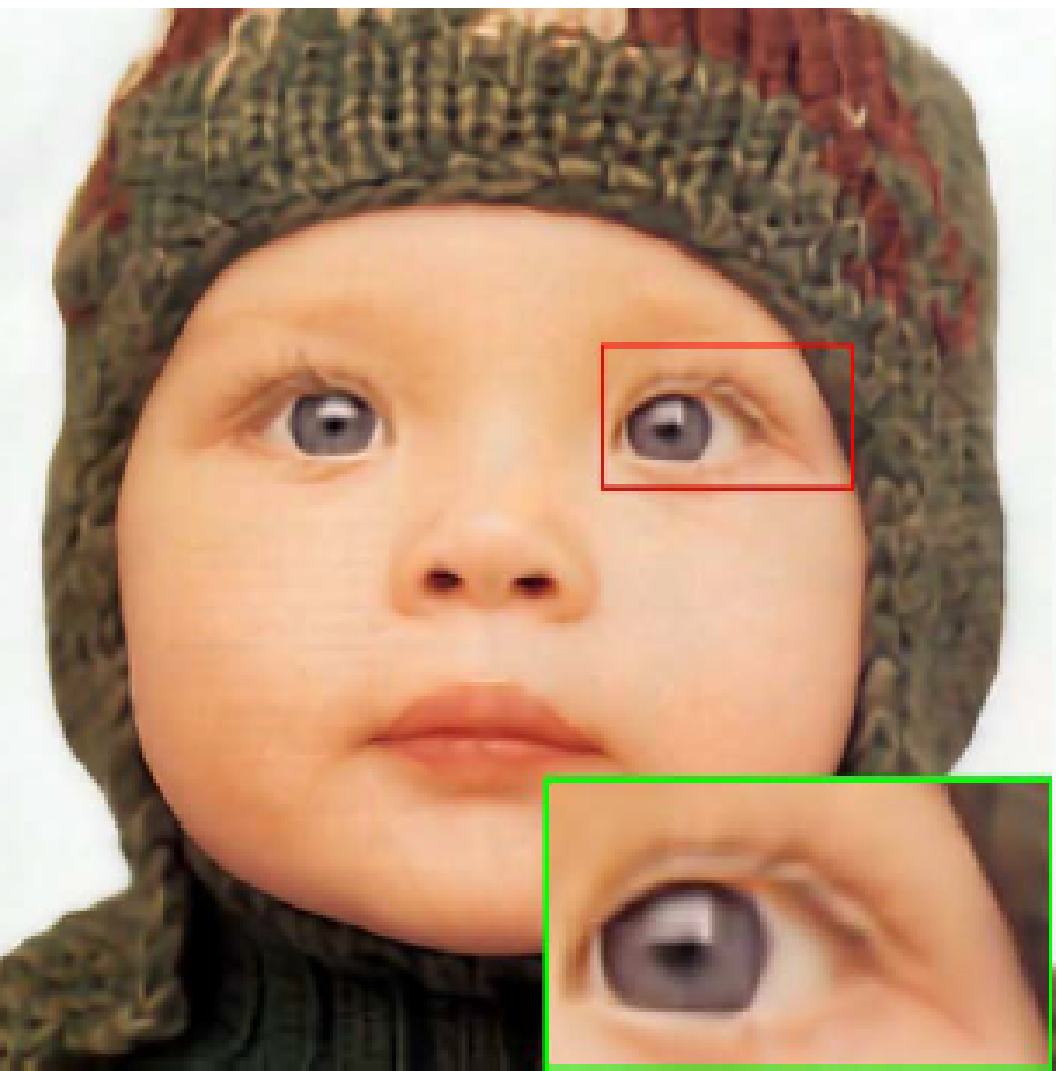}
       \caption{{\cite{yang2013fast}}}
    \end{subfigure}%
   \begin{subfigure}[b]{0.17\textwidth}
   \centering
        \includegraphics[width=0.75in]{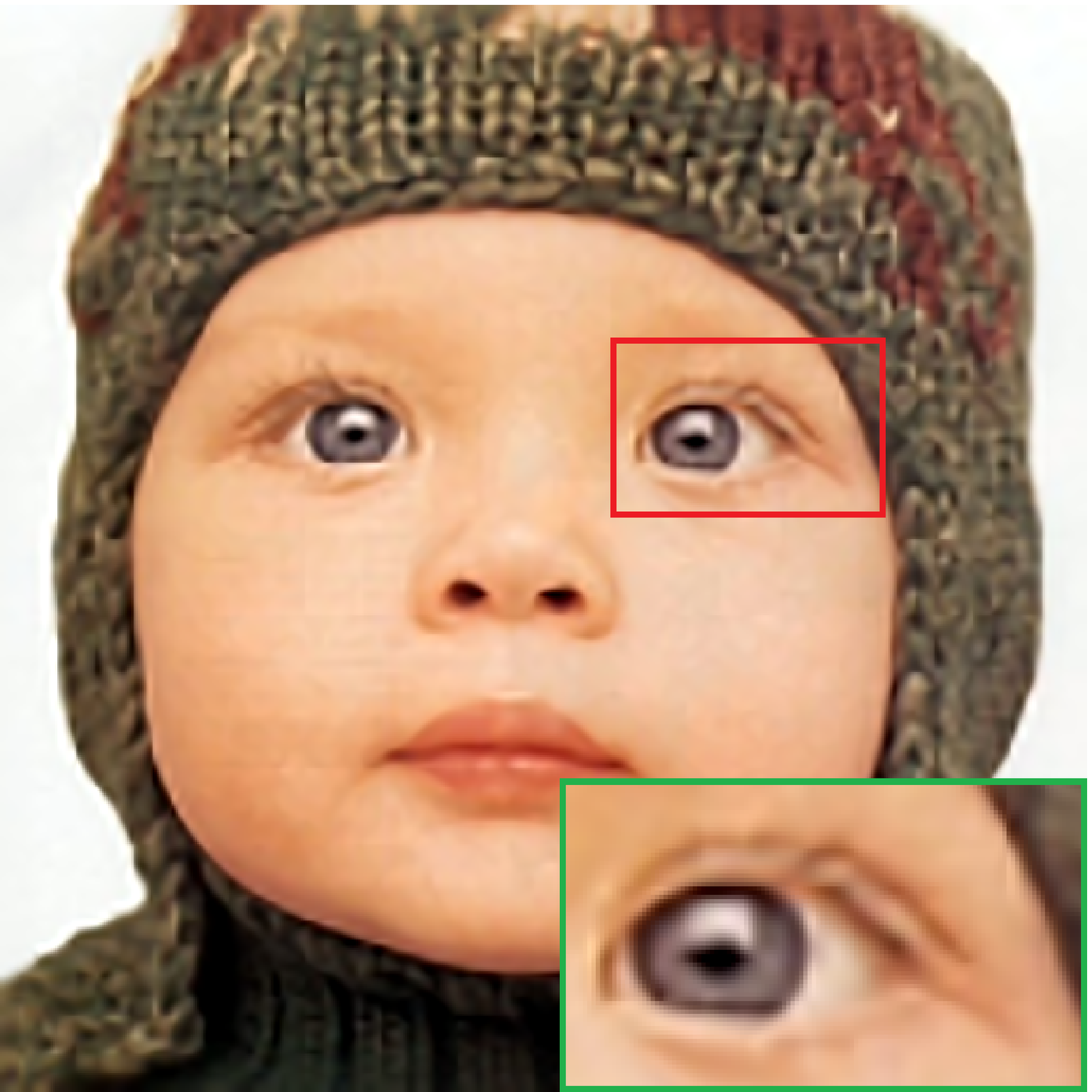}
        \caption{{Ours}}
   \end{subfigure}%
    \caption{SR of image \lq child\rq , \lq chip\rq .}
    \label{fig:comparison of SR image}
\end{figure}

In this section some visual results and comparisons with other methods are presented. We compare our approach with the recent state-of-the-art algorithms which are explained in the following articles \cite{xian2016single,yang2013fast1,shan2008fast,sun2011gradient} in terms of commonly used super-resolution test examples which can be seen in the Berkeley dataset BSDS500. For methods based on gradient information we select the algorithms presented in articles \cite{sun2011gradient,xian2016single} for comparison. For methods based on example-based information we chose algorithms presented in the articles \cite{glasner2009super,yang2013fast,yang2013fast1}for comparison. For methods based on reconstruction-based techniques, we used the algorithms in the articles \cite{freedman2011image,shan2008fast} for comparison. For visual quality comparison, most SR methods can produce artifacts-free images, however they fail to provide sufficient texture information. The artifacts-free image standard is what makes the images look very natural. The texture feature preserves the main structure of image, which has been regarded as the an essential characteristic of the image when used in image analysis and recognition. In practical applications, it is important for the image to display both texture and image naturalness. Therefore, in the following comparisons, we focus on the image similarity and texture similarity. \\
For image similarity, we compare different approaches using image samples of a \lq child\rq and a \lq chip\rq which are magnified by four. The results show that our method can produce more realistic texture details with minimum artifacts as compared with other methods. Moreover, in our synthetic high-resolution image, both the edge and texture are clear. The articles \cite{yang2013fast,yang2012coupled,freedman2011image} represent recent cutting-edge SR techniques. These methods were able to produce clear structure, however they also generate some blur in the richly textured region. Article \cite{glasner2009super} can recover more detail from the low-resolution image, but the results always cause over-artifacts as seen in Fig. \ref{fig:comparison of SR image}.
To measure the reconstruction error from the high-resolution synthetic image, we conduct quantitative analysis with bicubic and gradient profile prior methods in term of RMSE as better explained in articles \cite{wang2004image,yang2014single,sheikh2005information}. Table \ref{my-label} illustrates the results on high-resolution image. While other studies only magnified the test sample by three for simplicity, our study magnifies it by four while being able to produce better quality images. The table shows that our method outperforms the other methods by displaying a lower reconstruction error.\\
%\egin{figure}[htbp]
%  \centering
%   \begin{subfigure}[b]{0.16\textwidth}
%   \centering
%        \includegraphics[width=0.75in]{Figures/flower_1}
%    \end{subfigure}%
%     \centering
%   \begin{subfigure}[b]{0.16\textwidth}
%   \centering
%        \includegraphics[width=0.75in]{Figures/mushroom_1}
%    \end{subfigure}%
%   \begin{subfigure}[b]{0.16\textwidth}
%   \centering
%        \includegraphics[width=0.75in]{Figures/starfish_1}
%   \end{subfigure}%
%    \begin{subfigure}[b]{0.16\textwidth}
%   \centering
%        \includegraphics[height=0.75in,angle = 90]{Figures/girl_1}
%
%   \end{subfigure}%
%     \centering
%   \begin{subfigure}[b]{0.16\textwidth}
%   \centering
%        \includegraphics[height=0.75in,angle = 90]{Figures/lady_1}
%
%    \end{subfigure}%
%   \begin{subfigure}[b]{0.1\textwidth}
%   \centering
%        \includegraphics[height=0.5in]{Figures/lena_1}
%
%   \end{subfigure}%
%    \begin{subfigure}[b]{0.1\textwidth}
%   \centering
%        \includegraphics[height=0.5in]{Figures/kid_1}
%   \end{subfigure}%
%    \caption{Original test image.}
%    \label{fig:Original test image}
%\end{figur}

\begin{table}[H]
\centering
\caption{SR method quality measurement - RMSE}
\label{my-label}
\begin{tabular}{cccc}
\hline
Test image & Lena &Lady &Starfish  \\ \hline
bicubic       & 8.8     & 11.3             & 12.6     \\
gradient profile\cite{sun2011gradient}       & 7.8     & 9.5              &  11.5     \\
Ours   & \textbf{7.0707}    & \textbf{9.3883}            & \textbf{9.9912}     \\ \hline
\end{tabular}
\end{table}
Another test is conducted to analyze the similarity in term of  structural similarity (SSIM) as explained in articles \cite{wang2004image,yang2014single,sheikh2005information}. We chose four sample pictures consisting of a \lq child\rq, a \lq mushroom\rq, a \lq flower\rq and a \lq girl\rq in order to test patches similarity. As seen in Table \ref{my-label1}, we are able to show that the test results prove that our method outperforms most of the other methods that also display high image similarity.
\begin{table}[H]
\centering
\caption{SR method quality measurement - SSIM}
\label{my-label1}
\begin{tabular}{ccccc}
\hline
Test image & Child   & Mushroom   & Flower   & Girl    \\ \hline
Shan\cite{shan2008fast}        & \textbf{0.9183} & 0.8342 & 0.8713 & 0.8911 \\
Yang\cite{yang2013fast1}   & 0.9164 & 0.8542 & 0.8691 & 0.8891 \\
Xian\cite{xian2016single}     & 0.8906 & 0.7885 & 0.8332 & 0.8998 \\
Ours      & 0.8973 & \textbf{0.8618} & \textbf{0.8744} & \textbf{0.9077} \\ \hline
\end{tabular}
\end{table}
\subsection{Texture similarity}
\begin{figure}[htbp]
  \centering
  \begin{subfigure}[b]{0.32\textwidth}
   \centering
        \includegraphics[width=1.5in]{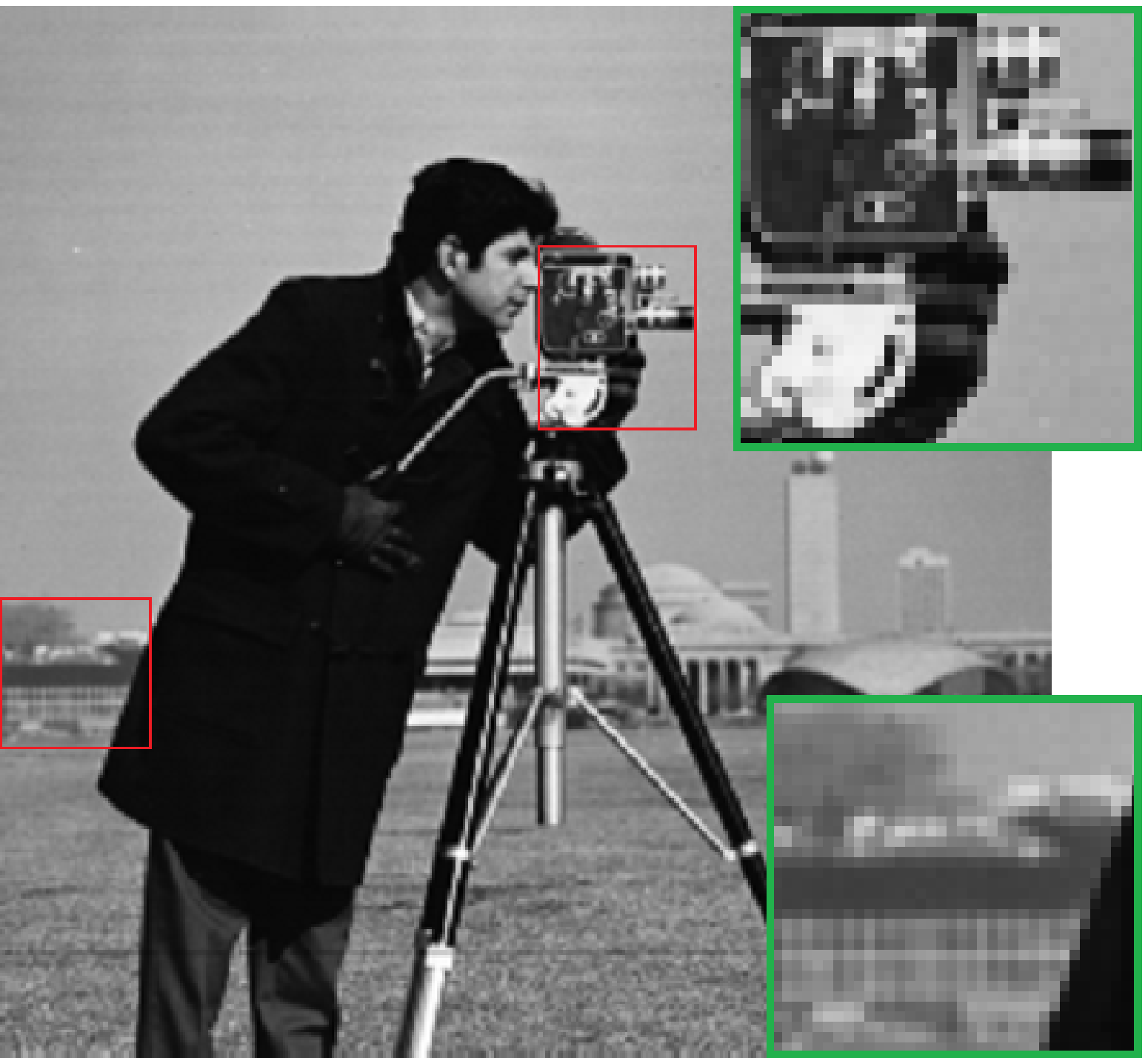}
    \end{subfigure}%
   \begin{subfigure}[b]{0.32\textwidth}
   \centering
        \includegraphics[width=1.5in]{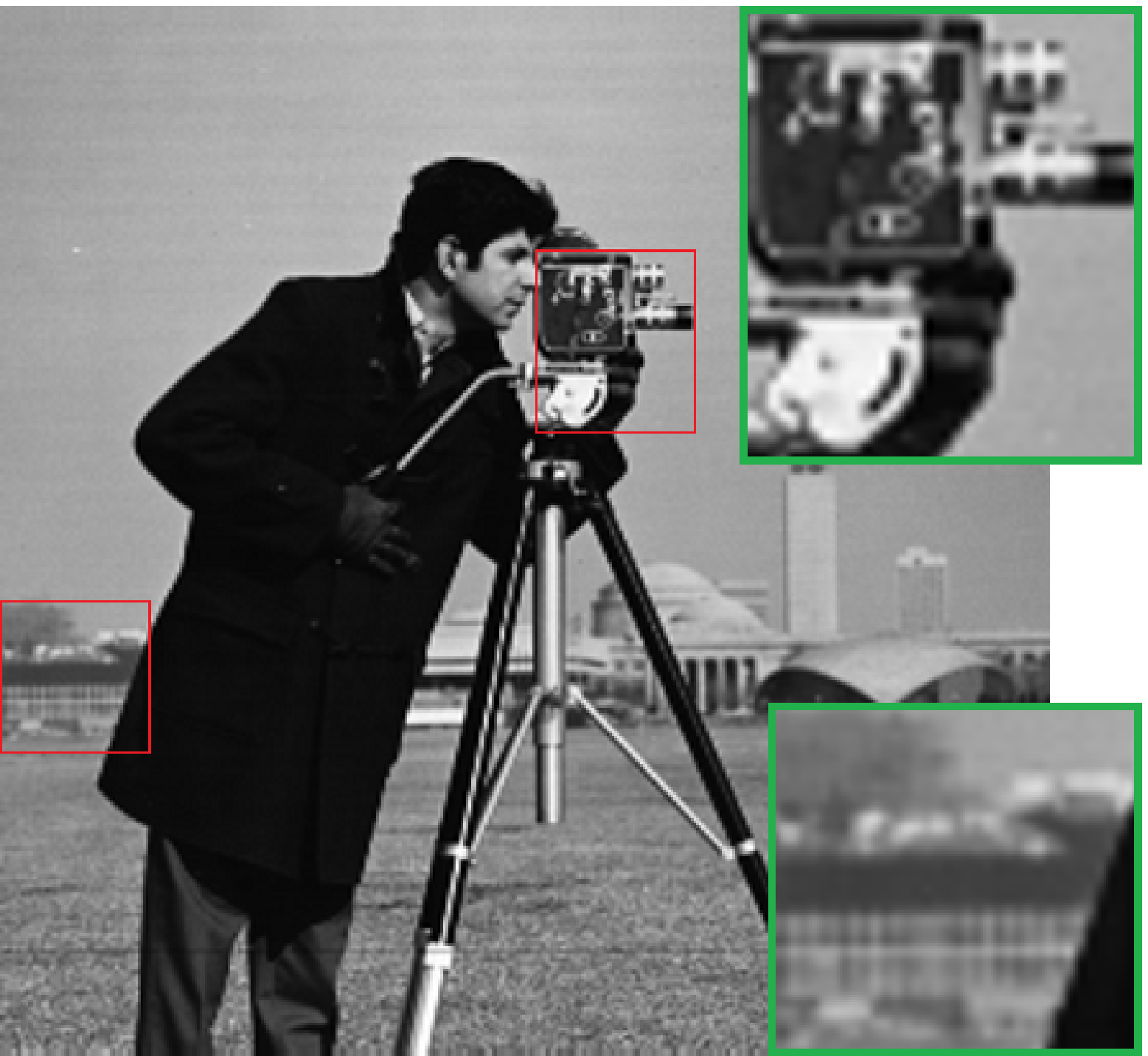}
   \end{subfigure}%
     \centering
   \begin{subfigure}[b]{0.32\textwidth}
   \centering
        \includegraphics[width=1.5in]{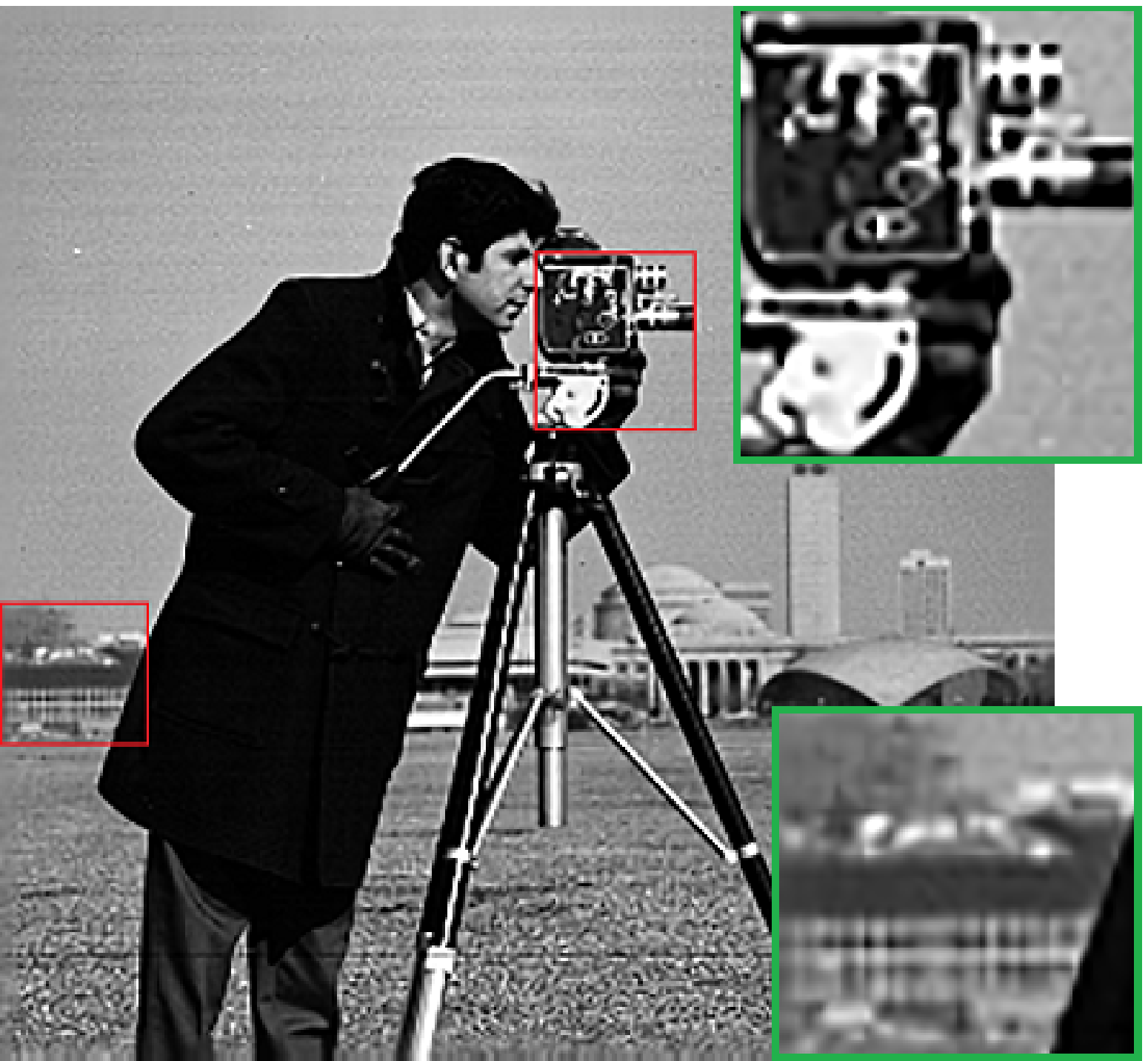}
    \end{subfigure}%
    \\
    \begin{subfigure}[b]{0.32\textwidth}
   \centering
        \includegraphics[width=1.5in]{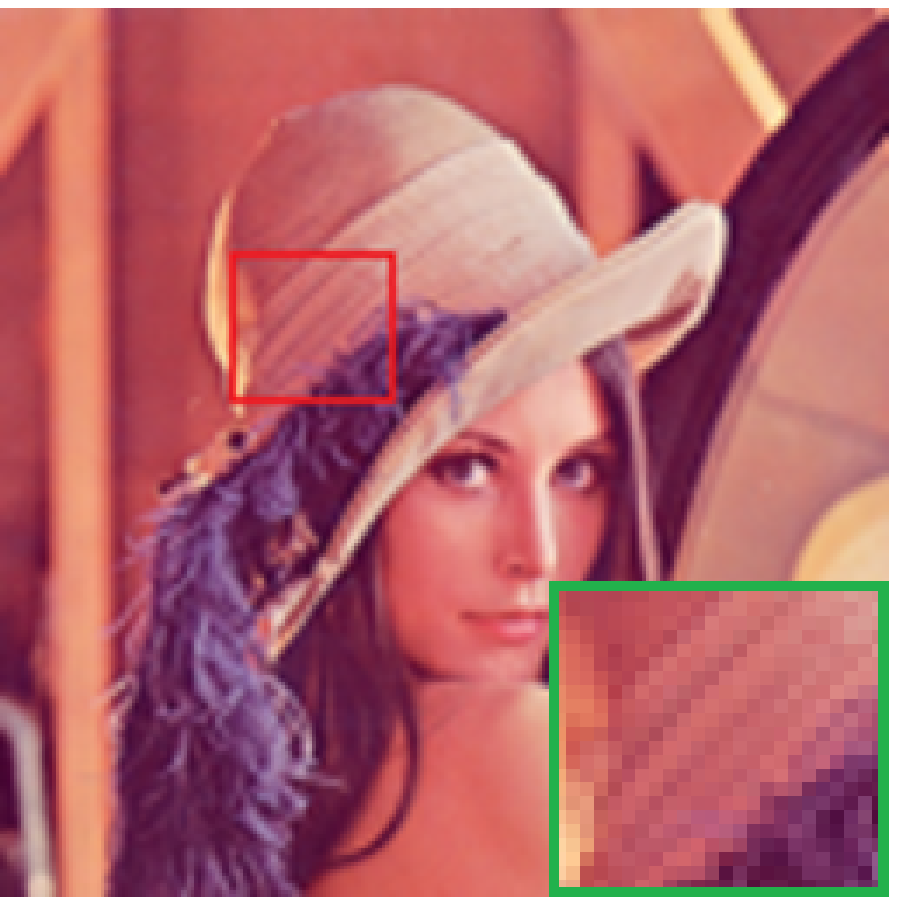}
    \end{subfigure}%
   \begin{subfigure}[b]{0.32\textwidth}
   \centering
        \includegraphics[width=1.5in]{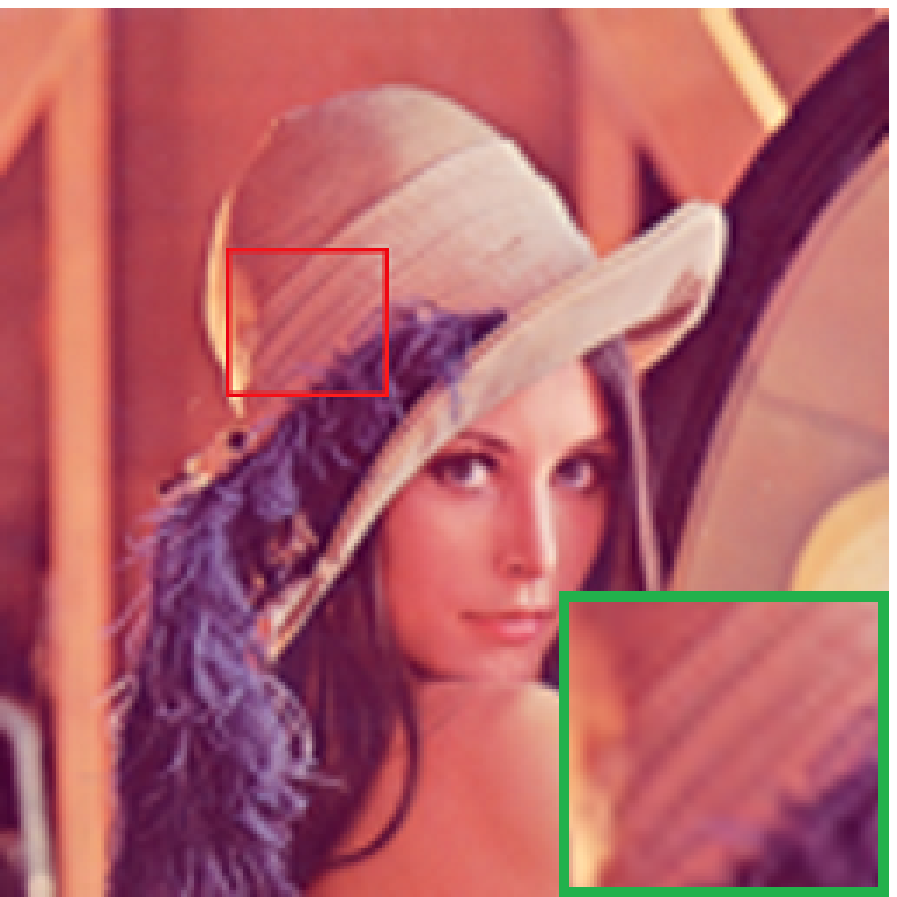}
   \end{subfigure}%
     \centering
   \begin{subfigure}[b]{0.32\textwidth}
   \centering
        \includegraphics[width=1.5in]{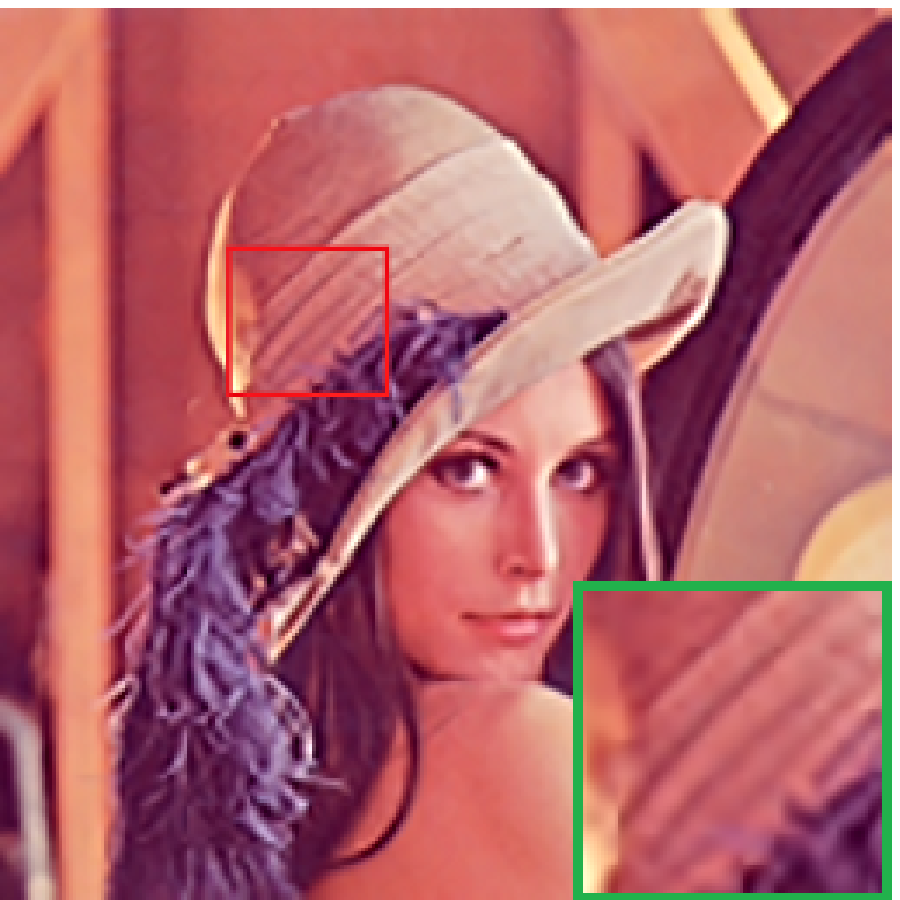}
    \end{subfigure}%
    \\
    %\begin{subfigure}[b]{0.32\textwidth}
%   \centering
%        \includegraphics[width=2in]{Figures/coins1}
%    \end{subfigure}%
%   \begin{subfigure}[b]{0.32\textwidth}
%   \centering
%        \includegraphics[width=2in]{Figures/coins_bi1}
%   \end{subfigure}%
%     \centering
%   \begin{subfigure}[b]{0.32\textwidth}
%   \centering
%        \includegraphics[width=2in]{Figures/coins-4}
%    \end{subfigure}%
%    \\
   \begin{subfigure}[b]{0.32\textwidth}
   \centering
        \includegraphics[width=1.5in]{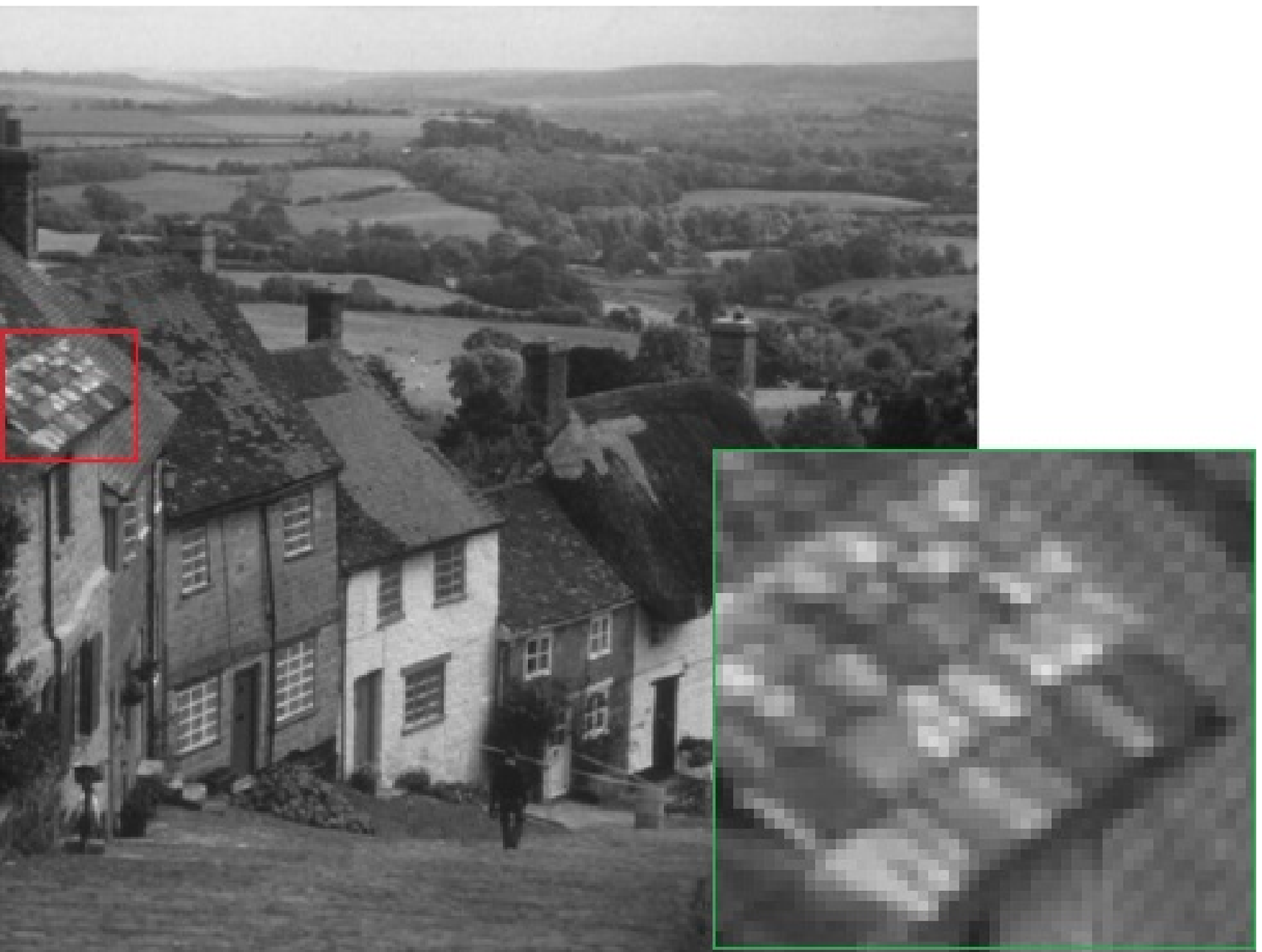}
    \end{subfigure}%
   \begin{subfigure}[b]{0.32\textwidth}
   \centering
        \includegraphics[width=1.5in]{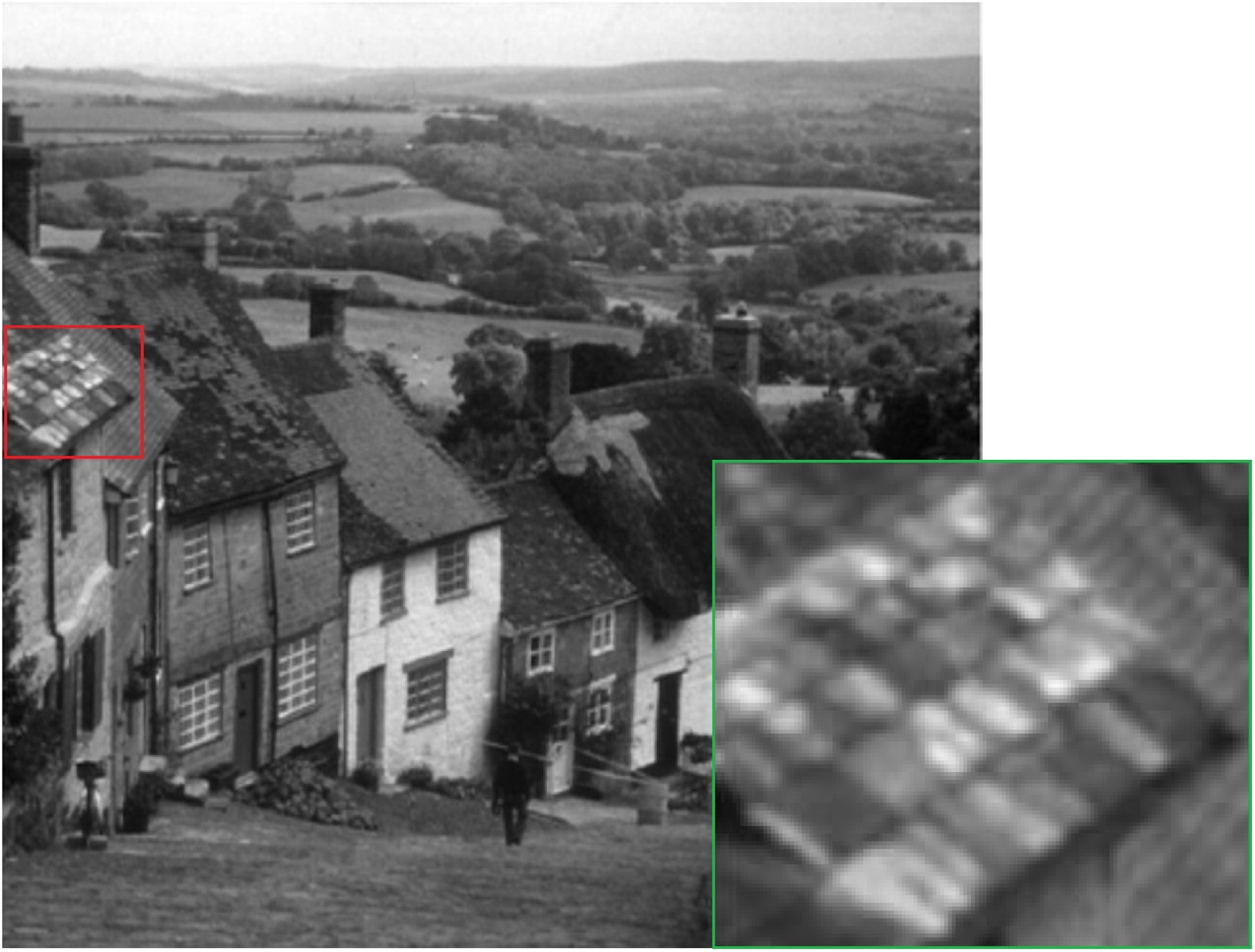}
   \end{subfigure}%
     \centering
   \begin{subfigure}[b]{0.32\textwidth}
   \centering
        \includegraphics[width=1.5in]{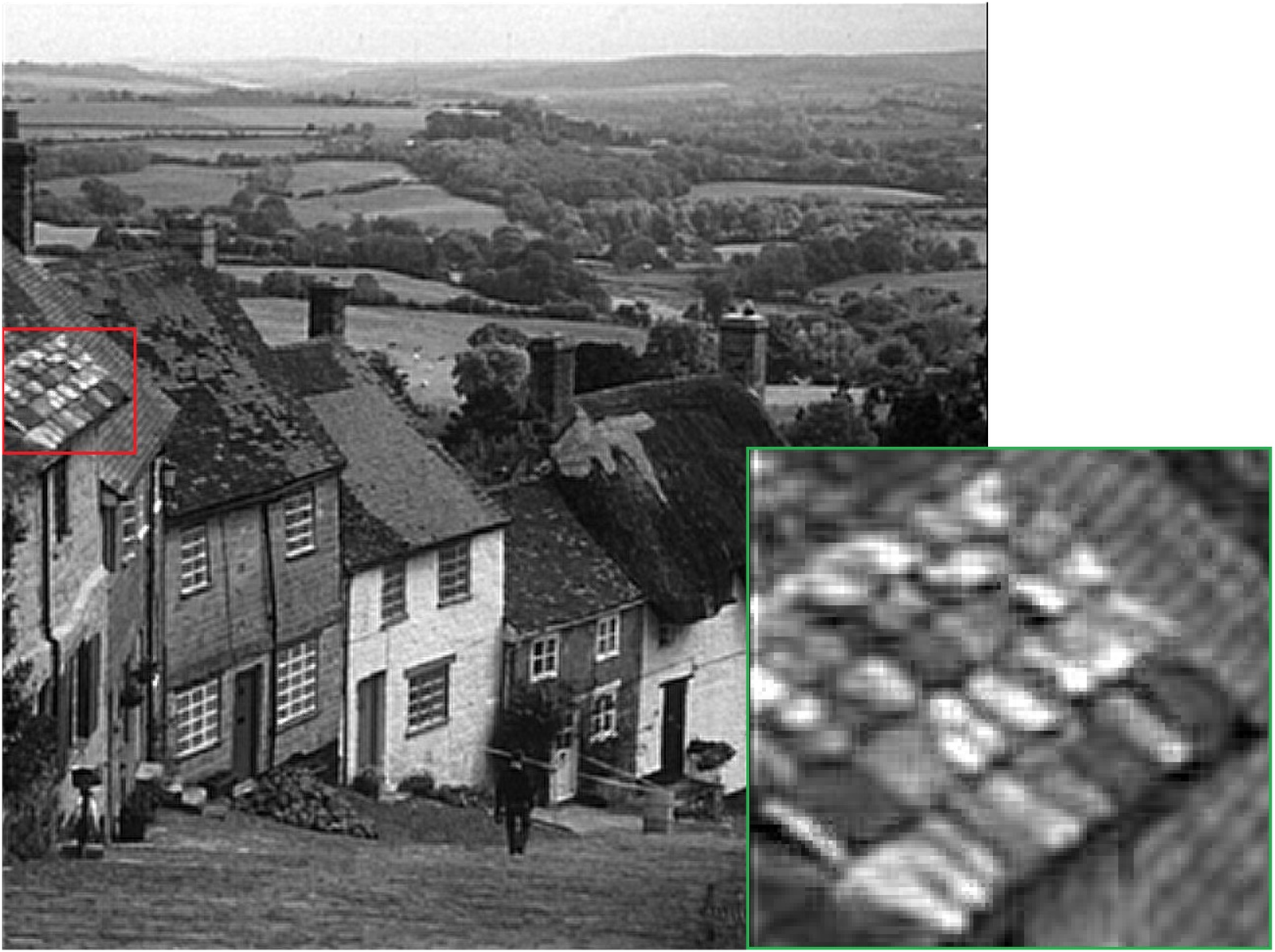}
    \end{subfigure}%
    \\
    \begin{subfigure}[b]{0.32\textwidth}
    \centering
        \includegraphics[width=1.5in]{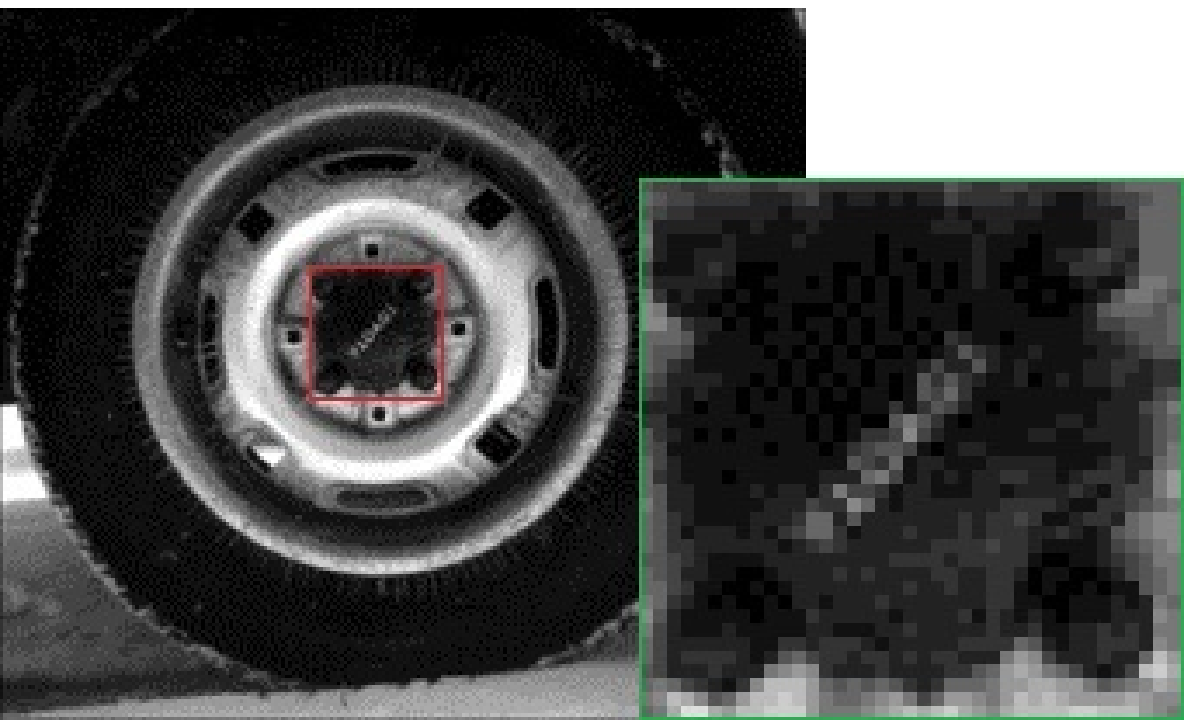}
    \end{subfigure}%
   \begin{subfigure}[b]{0.32\textwidth}
   \centering
        \includegraphics[width=1.5in]{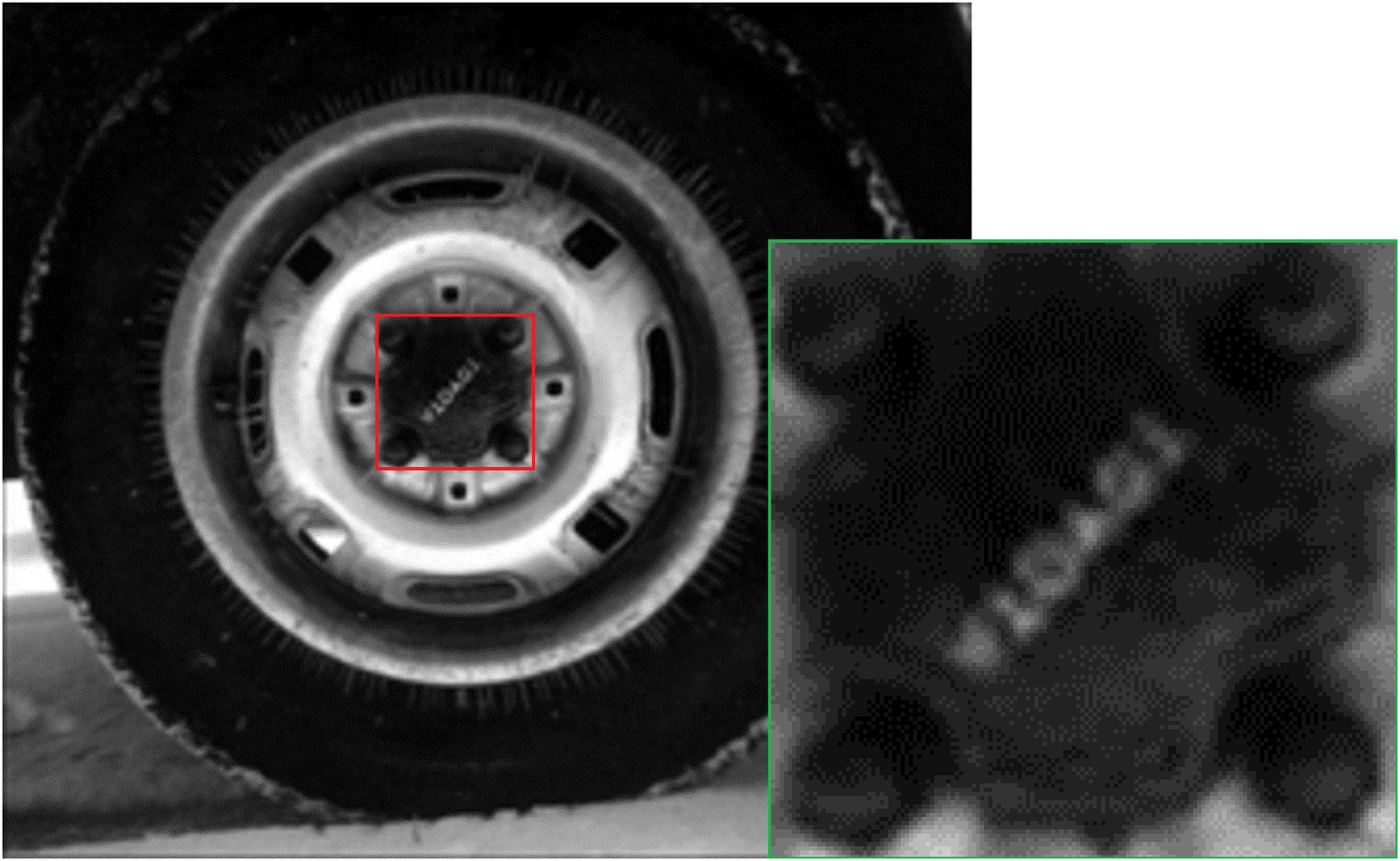}
   \end{subfigure}%
     \centering
   \begin{subfigure}[b]{0.32\textwidth}
   \centering
        \includegraphics[width=1.5in]{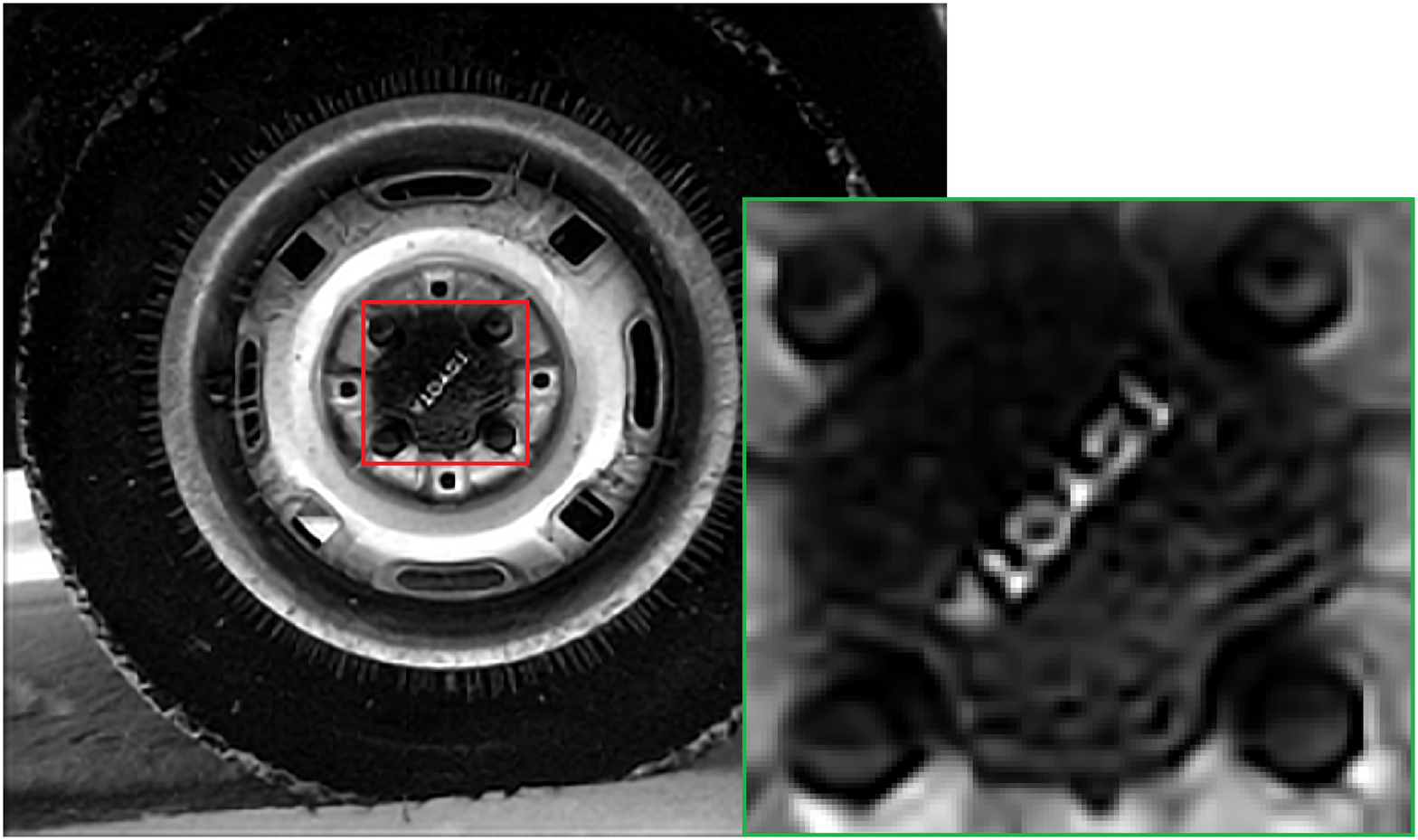}
    \end{subfigure}%
    \\
    \begin{subfigure}[b]{0.32\textwidth}
   \centering
        \includegraphics[width=1.5in]{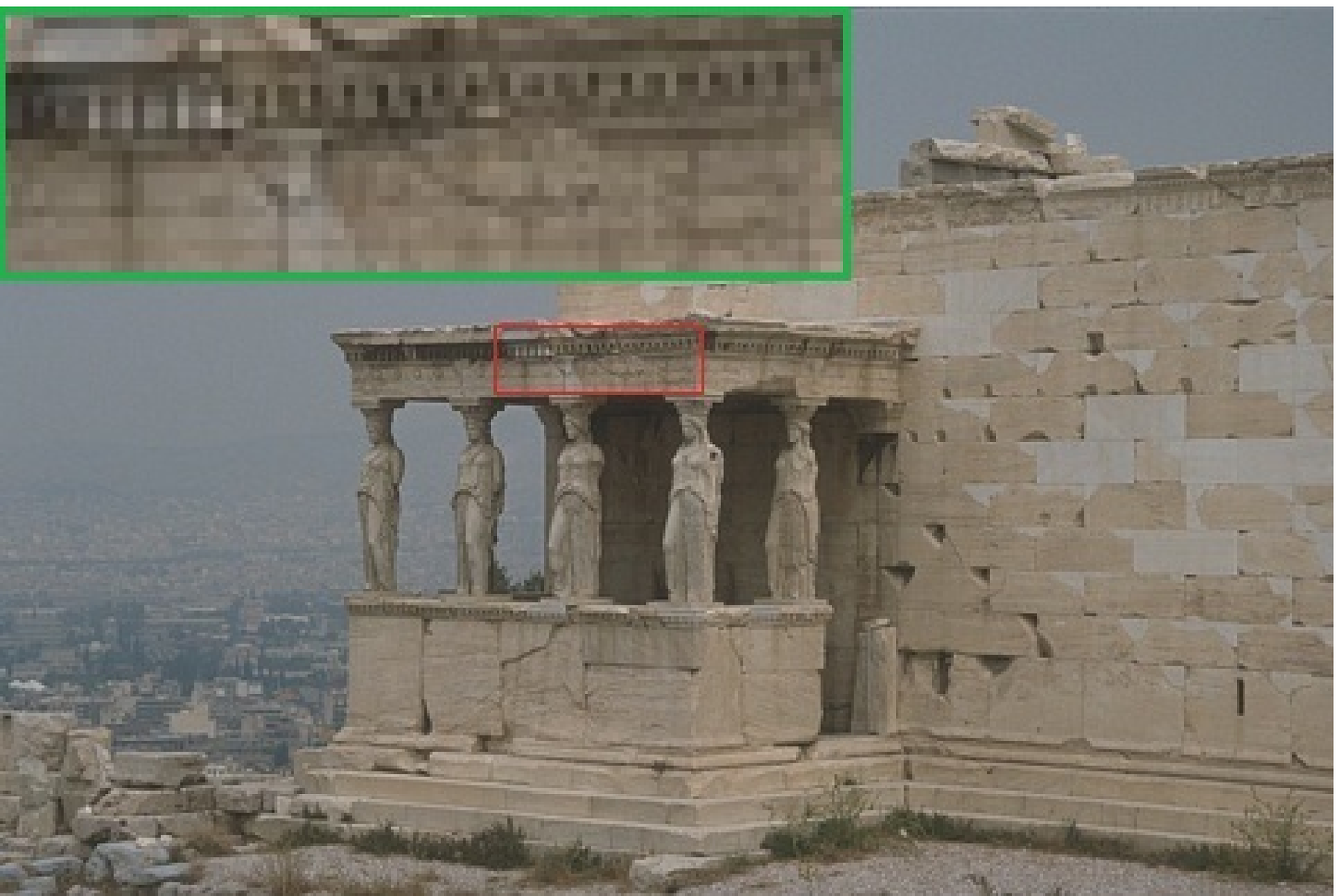}
        \caption{{\tt Original}}
    \end{subfigure}%
   \begin{subfigure}[b]{0.32\textwidth}
   \centering
        \includegraphics[width=1.5in]{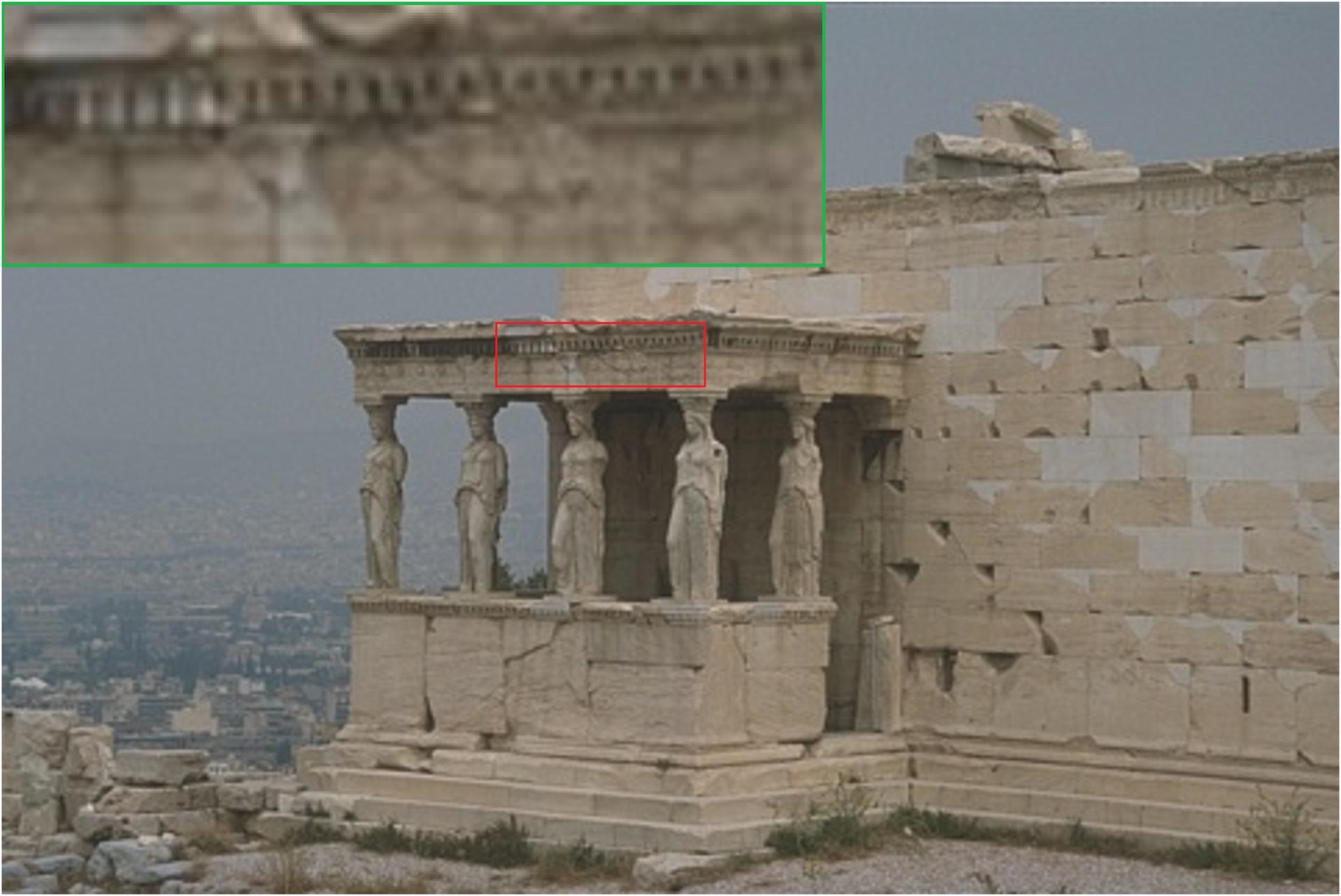}
        \caption{{\tt bicubic}}
   \end{subfigure}%
     \centering
   \begin{subfigure}[b]{0.32\textwidth}
   \centering
        \includegraphics[width=1.5in]{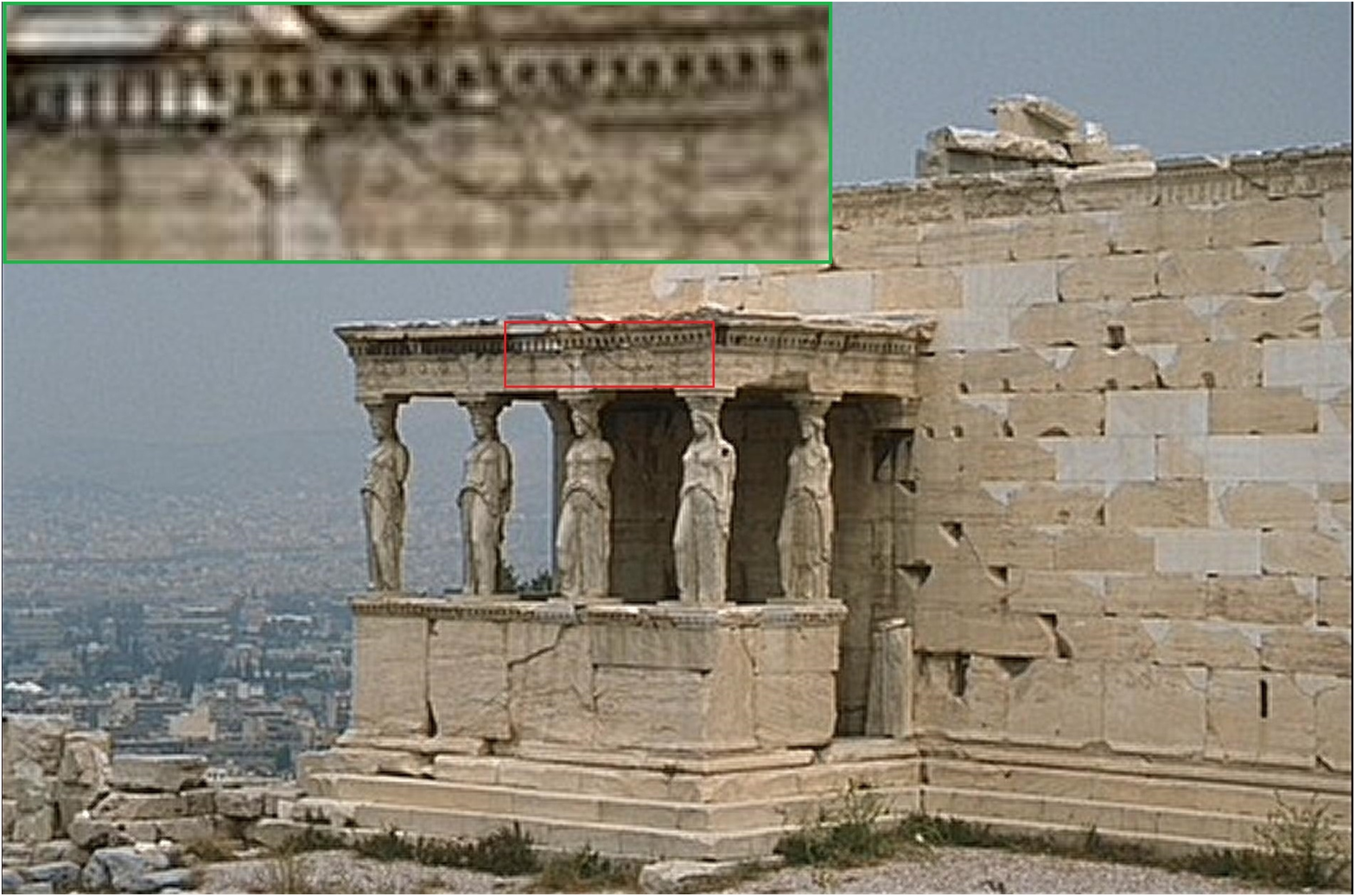}
        \caption{{\tt Ours}}
    \end{subfigure}%
    \caption{Comparison of the SR method to show texture details $\times 4$.}
    \label{fig:Comparison of the SR method to show texture details}
\end{figure}

\begin{figure}[htbp]
  \centering
   \begin{subfigure}[b]{0.5\textwidth}
   \centering
        \includegraphics[width=2.2in]{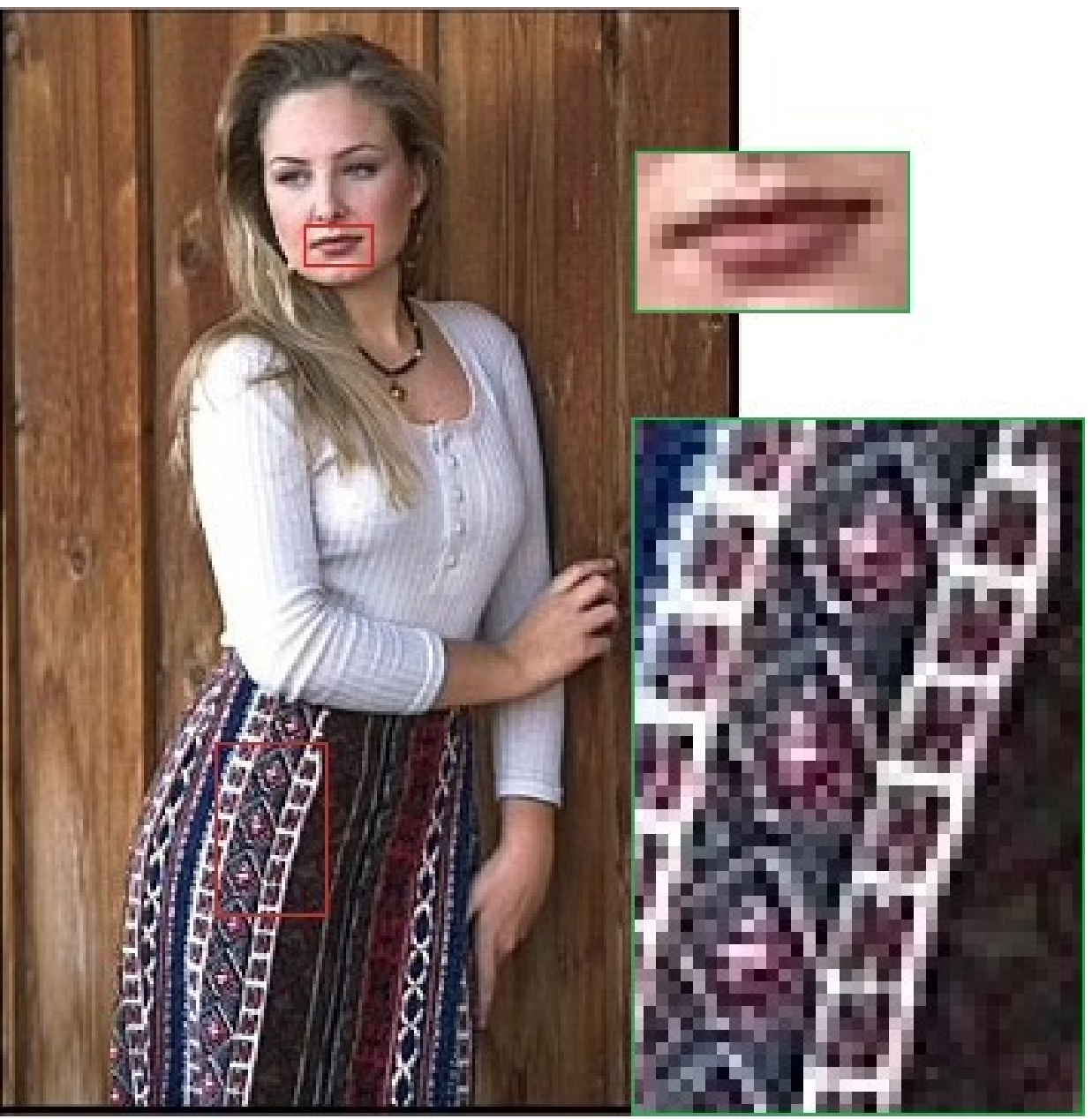}
        \caption{{\tt Original}}
    \end{subfigure}%
   \begin{subfigure}[b]{0.5\textwidth}
   \centering
        \includegraphics[width=2.2in]{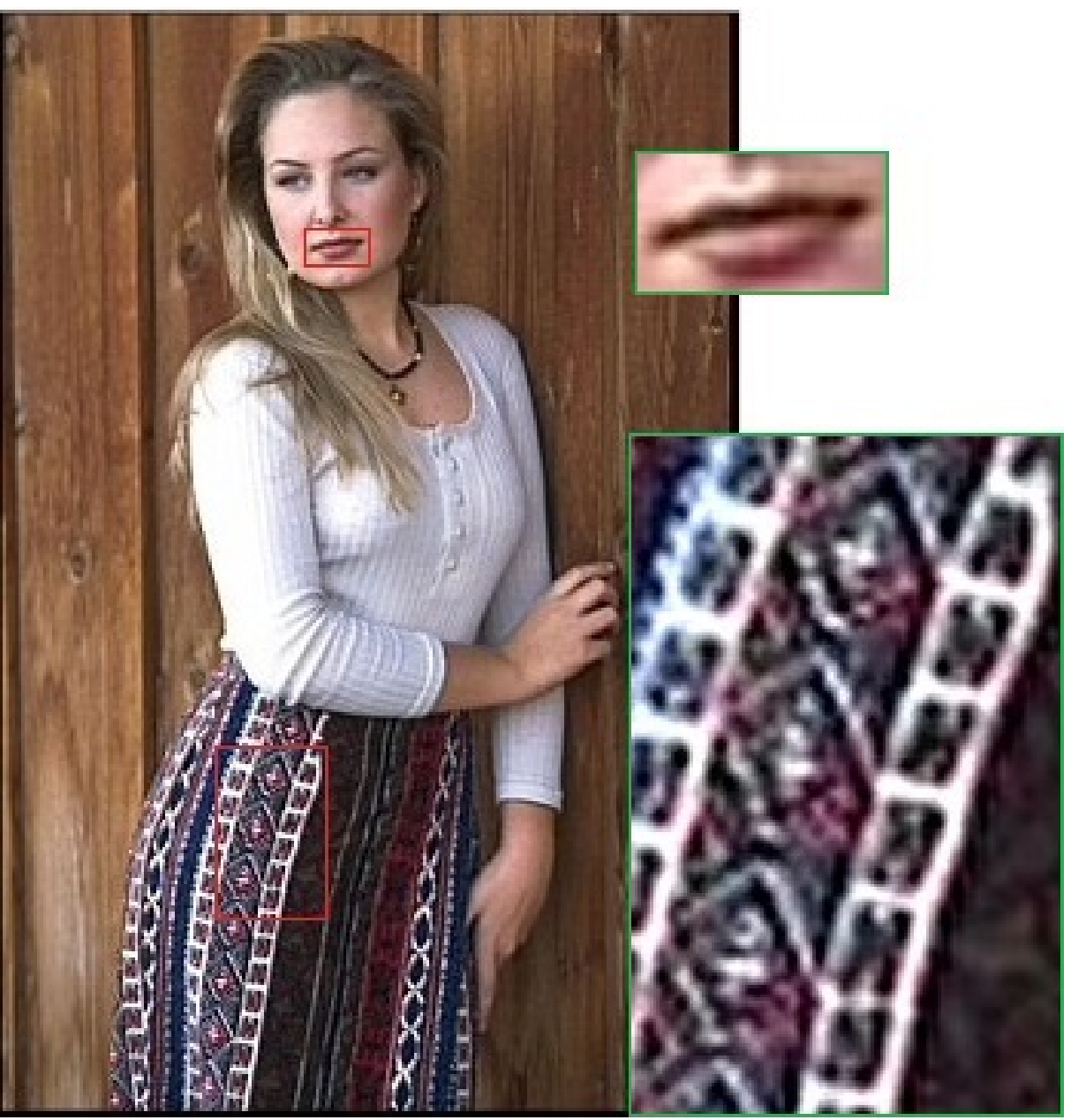}
        \caption{{\tt Ours}}
   \end{subfigure}%
    \caption{Illustration the texture recovery in high-resolution $\times 8$. This figure is better viewed on screen with high-resolution display due to the page size limit.}
    \label{fig:Illustration the texture recovery in high-resolution}
\end{figure}

\begin{figure}[htbp]
  \centering
   \begin{subfigure}[b]{0.45\textwidth}
   \centering
        \includegraphics[width=2.2in]{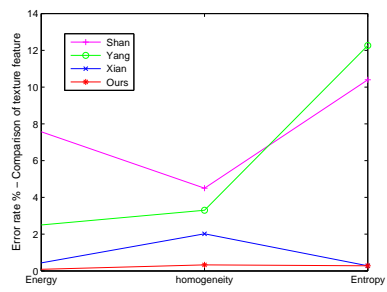}
        \caption{{\tt Kid }}
    \end{subfigure}%
   \begin{subfigure}[b]{0.45\textwidth}
   \centering
        \includegraphics[width=2.2in]{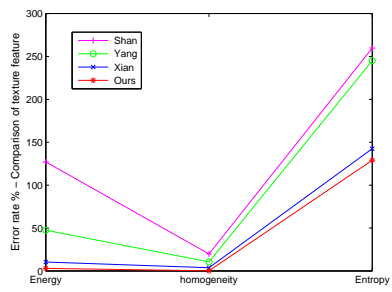}
        \caption{{\tt Mushroom }}
   \end{subfigure}%
     \centering
     \\
   \begin{subfigure}[b]{0.45\textwidth}
   \centering
        \includegraphics[width=2.2in]{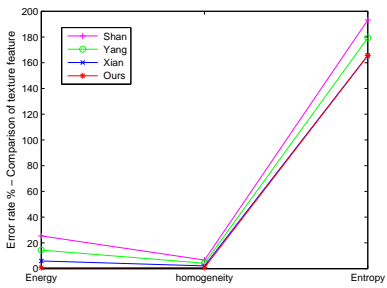}
        \caption{{\tt Flower }}
    \end{subfigure}%
    \begin{subfigure}[b]{0.45\textwidth}
   \centering
        \includegraphics[width=2.2in]{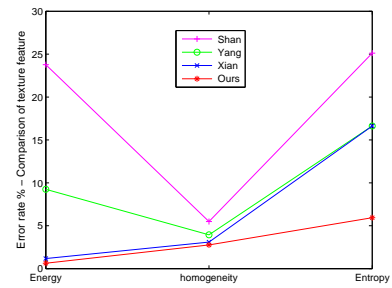}
        \caption{{\tt Girl }}
    \end{subfigure}%
    \caption{Comparison of the texture feature.}
    \label{fig:Comparison of the texture features}
\end{figure}

Image texture is a set of features designed to quantify an image. Image texture gives us information about the spatial arrangement of color or intensities of an image which can be artificially created or measured in natural scenes. Image textures represent the main structure information that can be used to help in segmentation or classification. Our SR method can generate enough texture detail from low-resolution images that the synthetic high-resolution images look very clear visually. We chose five image samples consisting of a \lq cameraman\rq, a woman \lq lena\rq, a \lq house\rq, a \lq tire \rq and a \lq statue \rq which are all magnified by four in order to test the texture details. From Fig. \ref{fig:Comparison of the SR method to show texture details}, we can see that our methods were able to synthesize a more clear texture than any other method. Our high-resolution images have sharp and clean structural texture and edges. Our high-resolution images are also demonstrated in Fig. \ref{fig:example}. We use image of a \lq lady\rq which is magnified by eight in order to illustrate the superiority of textures and edge recovery under large-scale SR. In Fig. \ref{fig:Illustration the texture recovery in high-resolution}, the mouse and the skirt have been recovered to display a clear texture. This figure is better viewed on screen with a high-resolution display due to the page size limit.\\
Texture similarity test is conducted to analyze the error in terms of following texture features: Energy, Homogeneity, and Entropy. We chose four pictures of a \lq child\rq, a \lq mushroom\rq, a \lq flower \rq and a \lq girl \rq in order to test patch texture similarity. The test results show that our method outperforms all of other methods with high texture similarity which can be seen in Tab \ref{my-label2}. Fig. \ref{fig:Comparison of the texture features} shows the comparison of the error rate that we produced on the four sample images versus error rates of the other methods.
% \usepackage{multirow}
\begin{table}[H]
\centering
\caption{SR method quality measurement - texture similarity}
\label{my-label2}
\begin{tabular}{ccccccc}
\hline
Test image                &             & Methods  &        &        &        &        \\\cline{2-7}
                          &             & Original & Shan\cite{shan2008fast}   & Yang\cite{yang2013fast1}   & Xian\cite{xian2016single}   & Ours   \\ \hline
\multirow{3}{*}{Child}      & Energy      & 0.2485   & 0.2673 & 0.2547 & 0.2496 & \textbf{0.2483} \\
                          & Homogeneity & 0.9186   & 0.9599 & 0.9489 & 0.9372 & \textbf{0.9216} \\
                          & Entropy     & 1        & 0.8960 & 0.8774 & 0.9972 & \textbf{0.9972} \\
\multirow{3}{*}{Mushroom} & Energy      & 0.0472   & 0.1072 & 0.0696 & 0.0521 & \textbf{0.0458} \\
                          & Homogeneity & 0.7068   & 0.8466 & 0.7816 & 0.7336 & \textbf{0.7078} \\
                          & Entropy     & 0.2730   & 0.9823 & 0.9422 & 0.6620 & \textbf{0.6253} \\
\multirow{3}{*}{Flower}   & Energy      & 0.1477   & 0.1852 & 0.1689 & 0.1565 & \textbf{0.1466} \\
                          & Homogeneity & 0.8852   & 0.9443 & 0.9222 & 0.9040 & \textbf{0.8781} \\
                          & Entropy     & 0.3373   & 0.9887 & 0.9422 & 0.8960 & \textbf{0.8960}  \\
\multirow{3}{*}{Girl}     & Energy      & 0.6483   & 0.8024 & 0.7083 & 0.6559 & \textbf{0.6442} \\
                          & Homogeneity & 0.9311   & 0.9822 & 0.9678 & 0.9599 & \textbf{0.9568} \\
                          & Entropy     & 0.8351   & 0.6253 & 0.6962 & 0.6962 & \textbf{0.8351} \\\hline
\end{tabular}
\end{table}

\section{Conclusion}
\label{sec:Conclusion}
In this paper, we have presented a new single image super-resolution method based on adaptive fractional-order gradient interpolation and reconstruction. As illustrated by the experimental results, our method is able to synthesize sharp edges while preserving texture information. By using the fractional-order gradient interpolation we are able to ensure sharp and clear edges and more texture details, and by adopting the minimum energy function to optimize the final high-resolution image we are able to ensure the image and texture similarity. The proposed approach is robust under multi-scale super-resolution condition and could generate excellent high-resolution images. For an image of size 128x128 we set the SR scale to four and with the use of a computer with a graphics processing unit (GPU), and with the use of Matlab 2016 we are able to see that our algorithm produces a cost time of approximately 1-2 seconds.

\section*{Acknowledgements}
The work was partially supported by the Liaoning Province Education Project No. L2014082 and China Scholarship Council. Acknowledgement goes to Twinkle Mistry for final revisions of this paper.

\section*{References}

%\bibliography{mybibfile}
\bibliography{Image_Processing_bibfile}
\end{document}